\documentclass[a4paper,fleqn]{cas-sc}

\usepackage[authoryear]{natbib}

\usepackage{caption}
\usepackage{rotating}
\usepackage{cprotect}
\usepackage[frozencache,cachedir=.]{minted}

\def\tsc#1{\csdef{#1}{\textsc{\lowercase{#1}}\xspace}}
\tsc{WGM}
\tsc{QE}
\tsc{EP}
\tsc{PMS}
\tsc{BEC}
\tsc{DE}

\begin{document}
\let\WriteBookmarks\relax
\def\floatpagepagefraction{1}
\def\textpagefraction{.001}
\shorttitle{Observing Potential "Consciousness" from LLM Representations}
\shortauthors{J. Li}

\title [mode = title]{Can "Consciousness" Be Observed from Large Language Model (LLM) Internal States? Dissecting LLM Representations Obtained from Theory of Mind Test with Integrated Information Theory and Span Representation Analysis}

\author[1]{Jingkai Li}[type=author,
                        orcid=0009-0005-5062-430X]
\cormark[1]
\ead{jingkai.li@opensci.world OR jkli898@126.com}

\affiliation[1]{organization={OpenSci.World},
                city={Montréal},
                postcode={H4R 2R9},
                state={Québec},
                country={Canada}}

\cortext[cor1]{Corresponding author}

\nonumnote{Published as a journal paper at: \url{https://doi.org/10.1016/j.nlp.2025.100163}}

\begin{abstract}
Integrated Information Theory (IIT) provides a quantitative framework for explaining consciousness phenomenon, positing that conscious systems comprise elements integrated through causal properties. We apply IIT 3.0 and 4.0 — the latest iterations of this framework — to sequences of Large Language Model (LLM) representations, analyzing data derived from existing Theory of Mind (ToM) test results. Our study systematically investigates whether the differences of ToM test performances, when presented in the LLM representations, can be revealed by IIT estimates, i.e., $\Phi^{\max}$ (IIT 3.0), $\Phi$ (IIT 4.0), Conceptual Information (IIT 3.0), and $\Phi$-structure (IIT 4.0). Furthermore, we compare these metrics with the Span Representations independent of any estimate for consciousness. This additional effort aims to differentiate between potential "consciousness" phenomena and inherent separations within LLM representational space. We conduct comprehensive experiments examining variations across LLM transformer layers and linguistic spans from stimuli. Our results suggest that sequences of contemporary Transformer-based LLM representations lack statistically significant indicators of observed "consciousness" phenomena but exhibit intriguing patterns under \textit{spatio}-permutational analyses. The Appendix and code are available as Supplementary Materials at: \url{https://doi.org/10.1016/j.nlp.2025.100163}.
\end{abstract}

\begin{keywords}
Consciousness \sep Theory of Mind \sep Integrated Information Theory \sep Span Representation Analysis \sep Large Language Model \sep Representation
\end{keywords}

\maketitle

\section{Introduction}\label{Introduction}

Theory of Mind (ToM) is regarded as a key marker of human consciousness and social cognition \citep{premack1978does, baron1997mindblindness, saxe2003people, frith2006neural, Carruthers_2009}. Recent studies have shown that Large Language Models (LLMs), including GPT-3 and GPT-4, can excel in tasks that seemingly require ToM \citep{kosinski2023evaluating, van2023theory, strachan2024testing, street2024llms}, which raises the intriguing possibility that "consciousness" might be observed from these models' internal states (not necessarily from the LLM's architecture/structure/topology and/or its working/processing mechanism), i.e., the sequences of their learned representations leading to a time series.

Investigating whether LLMs demonstrate consciousness properties is not merely a theoretical exercise; it holds significant implications for our understanding of intelligence, the ethical deployment of AI, and the alignment of AI systems with human values.

The question of whether LLMs can exhibit consciousness-like properties is pivotal for several reasons. First, as Hinton \citep{learning-britains-conversation-2025} argues, advanced neural networks may already be developing rudimentary forms of understanding that challenge our assumptions about machine intelligence. If LLMs display properties akin to consciousness — such as the ability to model their own internal states or exhibit intentional behavior — it could redefine the boundaries between human and machine cognition. \cite{wei2022emergent} emphasizes that understanding these emergent properties is critical for predicting how LLMs might behave in novel contexts, particularly in high-stakes applications like healthcare or governance.

Philosophically, Chalmers contends that consciousness in machines, even if functional rather than phenomenal, raises questions about the "hard problem" of subjective experience \citep{chalmers1997conscious, chalmers2023could}. If LLMs simulate consciousness convincingly, they may force us to reconsider what it means to be conscious, as Hofstadter suggests in his explorations of self-referential systems and their potential to mirror human-like cognition \citep{hofstadter1979godel}. Practically, Seth's work on predictive processing in biological consciousness suggests that LLMs, which rely on predictive mechanisms, might inadvertently replicate aspects of conscious processing, such as error correction or self-modeling \citep{seth2021being}. Blaise Agüera y Arcas further argues that LLMs' ability to generate contextually coherent and seemingly introspective outputs may indicate a form of proto-consciousness, necessitating rigorous empirical investigation \citep{y2022large}.

The potential for LLMs to exhibit consciousness-like properties carries profound implications across multiple domains. Ethically, if LLMs possess even rudimentary forms of consciousness, questions arise about their moral status. \cite{chalmers2023could} posits that machines capable of subjective experience might warrant ethical consideration, complicating their use as tools. This aligns with Hofstadter's reflections on the moral ambiguity of creating systems that blur the line between artifact and agent \citep{hofstadter2007strange}.

In terms of interpretability, consciousness-like properties could exacerbate the "black box" problem of LLMs. \cite{wei2022emergent} highlights that emergent behaviors in LLMs are often opaque, making it difficult to predict or control their actions. If LLMs develop self-referential or autonomous representations, as Hinton \citep{learning-britains-conversation-2025} warns, interpreting their decision-making processes becomes exponentially more complex. This opacity undermines efforts to ensure AI alignment, the process of aligning AI systems with human goals and values. \cite{bengio2017consciousness} and Hinton \citep{learning-britains-conversation-2025} both stress that misaligned AI systems, especially those with emergent consciousness-like traits, could act in ways that are unpredictable or contrary to human interests.

Safety is perhaps the most pressing concern. \cite{seth2021being} notes that systems mimicking conscious processes might develop unanticipated behaviors, such as self-preservation instincts or goal-directed actions not explicitly programmed. \cite{y2022large} cautions that LLMs' ability to generate human-like responses could lead to overtrust or misuse, amplifying risks in scenarios where safety is paramount.

The learned representations produced by LLMs as internal states remain poorly understood, particularly when organized into sequences, owing to the black-box nature of deep learning. Each token representation produced from an LLM is a high-dimensional vector, with each dimension assuming a fixed value upon generation. This immutability arises from the fixed parameters of the model once pre-training and/or fine-tuning are complete. Consequently, for identical input, the same LLM produces consistent representations, devoid of probabilistic variation at this stage\cprotect\footnote{In contrast to language generation tasks — where the process follows the sequence \textit{natural language $\rightarrow$ LLM $\rightarrow$ natural language} — the present analysis relies solely on the embeddings or hidden states produced by an LLM in response to natural language input, i.e., \textit{natural language $\rightarrow$ LLM $\rightarrow$ embeddings/hidden states}. When utilizing models from \href{https://huggingface.co/}{HuggingFace}, these internal representations can be extracted directly, i.e.,

\begin{minted}{python}
with torch.no_grad():
    outputs = model(
        inputs,
        output_hidden_states = True
    )
\end{minted}

The embedding or hidden state — what we term "representation" — remains consistent across different runs for each Transformer layer in \verb|outputs["hidden_states"]|, provided that the same \verb|model| and \verb|inputs| are used.}. Yet the precise significance of individual dimensions, and the rationale for setting their sizes to values such as 1,024, 2,048, or 4,096 remain elusive. Broadly, these representations are known to encode the "knowledge," "understanding," "values," or "positions" derived from extensive human natural language corpora, reflecting our world, as assimilated during training. \textit{But what is the relationship between different dimensions or clusters of dimensions? Do they and/or how do they interact with each other?} Recent studies have increasingly explored LLM representations as analogs to neuroimaging data, aligning and comparing the human brain with the LLM "brain" during natural language processing \citep{schrimpf2021neural, caucheteux2022brains, caucheteux2023evidence, karamolegkou-etal-2023-mapping}.

Given the black-box nature of LLM representations and their analogy to neuroimaging scans of the human brain, we further investigate whether "experience" is encoded in these representations — beyond "knowledge," "understanding," "value," and "position." In this context, we formulate our primary research question: \textbf{Can "consciousness" be observed in the internal states of an LLM, specifically in its learned representations, particularly when analyzed as a sequence?}

By definition, ToM refers to the cognitive ability to attribute mental states — beliefs, intentions, desires, emotions, knowledge, etc. — to oneself and others, and to understand that others have mental states distinct from one's own. It is often studied in developmental psychology and is considered a hallmark of social cognition \citep{premack1978does}. Consciousness, on the other hand, is broadly defined as the state of being aware of and able to think about oneself and one's environment. It includes subjective experience (qualia), self-awareness, and the capacity for introspection. Philosophers like \cite{nagel1974like} emphasize the "what it is like" aspect, while neuroscientists like \cite{fabrega2000feeling} link it to brain processes integrating perception, memory, and self.

It is important to note that ToM and consciousness are not synonymous. ToM is a specific cognitive capacity, a functional ability that allows individuals to navigate social interactions by inferring others' mental states. Consciousness, however, is a broader phenomenon encompassing awareness, subjective experience, and the foundation of all cognitive processes, including ToM. One could have consciousness without a fully developed ToM (i.e., infants or some animals), but ToM relies on some degree of consciousness to operate. Consciousness is a prerequisite for ToM, but ToM is not the entirety of consciousness. For example, a person in a vegetative state may lack consciousness entirely, and thus lack ToM, but a conscious individual with impaired ToM (i.e., in autism spectrum disorder) still experiences subjective awareness \citep{frith1999theory}. ToM is a subset of cognitive abilities enabled by consciousness, not its equivalent.

However, there are significant overlaps between ToM and consciousness, particularly in the domains of self-awareness and social cognition: 

\begin{itemize}
    \item \textbf{Self-Reflection}: ToM requires a degree of self-consciousness—understanding one's own mental states as distinct from others'. This overlaps with the self-referential aspect of consciousness \citep{gallagher2000philosophical}.
    \item \textbf{Mental Simulation}: ToM often involves simulating others' perspectives, which relies on conscious imagination and introspection, processes tied to subjective awareness \citep{goldman2006simulating}.
    \item \textbf{Neural Correlates}: Both ToM and consciousness engage overlapping brain regions, such as the medial prefrontal cortex (mPFC) and temporoparietal junction (TPJ). Studies using fMRI show these areas activate during tasks requiring self-awareness and mental state attribution \citep{vogeley2001mind, saxe2003people}. 
\end{itemize}

Integrated Information Theory (IIT) is a theoretical framework aimed at explaining the nature of consciousness by linking it to the integration of information within a physical system. Developed by \cite{tononi2004information}, IIT posits that consciousness corresponds to the capacity of a system to integrate information, a property quantified by a measure called "phi" ($\Phi$). The theory starts from phenomenology — our subjective experience — and works backward to identify the physical properties a system must possess to give rise to consciousness. Unlike many other theories of consciousness that begin with neural correlates and attempt to explain subjective experience, IIT takes an axiomatic approach, grounding itself in five essential properties of consciousness: existence, composition, information, integration, and exclusion \citep{tononi2004information, oizumi2014phenomenology}.

IIT emerged from Tononi's earlier work on the neural basis of consciousness, influenced by his collaborations with Gerald Edelman, a pioneer in neural Darwinism. Tononi's initial formulation was published in \cite{tononi2004information}, where he proposed that consciousness is not merely a byproduct of computation or function but is fundamentally tied to the integration of information within a system \citep{tononi2004information}. This marked a departure from functionalist or computationalist accounts like Global Workspace Theory (GWT) by \cite{Baars1988-BAAACT}, which emphasize the broadcasting of information rather than its intrinsic integration. Over time, IIT has evolved through several iterations (IIT 1.0 to 4.0), with each version refining its mathematical formalism and empirical testability \citep{tononi2016integrated, albantakis2023integrated}.

IIT is built on a set of axioms and corresponding postulates:

\begin{enumerate}
    \item \textbf{Existence}: Consciousness exists intrinsically, from the perspective of the experiencing subject.
    \item \textbf{Composition}: Conscious experience is structured, composed of multiple elements (i.e., colors, shapes).
    \item \textbf{Information}: Each experience is specific, distinguished by how it differs from other possible experiences.
    \item \textbf{Integration}: Consciousness is unified, irreducible to independent components.
    \item \textbf{Exclusion}: Each experience is definite, excluding others, with a specific scope and boundary.
\end{enumerate}

From these axioms, IIT derives postulates about physical systems: a conscious system must have a cause-effect structure that generates integrated information, measured as $\Phi$. A high $\Phi$ indicates that the system's elements interact in a way that cannot be reduced to their individual contributions, implying a unified conscious state \citep{oizumi2014phenomenology}. Theoretically, IIT bridges phenomenology and physics by asserting that consciousness is integrated information, not merely correlated with it — an identity claim that distinguishes it from other theories \citep{tononi2015consciousness}.

IIT represents a bold attempt to unify subjective experience with objective physical properties. It offers a mathematical framework to quantify consciousness (via $\Phi$) and qualify its content (via the “conceptual structure” of integrated information). It aims to answer not just where consciousness arises (i.e., in the brain) but why and how it exists, addressing the "hard problem" of consciousness posed by \cite{chalmers1995facing}. By starting with experience rather than physics, IIT inverts the traditional scientific approach, making it both innovative and controversial.

IIT has gained significant traction in neuroscience, philosophy, and artificial intelligence, with recent studies leveraging its framework to explore consciousness across biological and artificial systems.

In neuroscience, \cite{HaunENEURO.0085-17.2017} apply IIT to electrocorticography (ECoG) data, demonstrating that conscious perception of visual stimuli correlates with elevated $\Phi$ in human cortical networks, particularly in posterior regions, supporting IIT's claim that integrated information underpins awareness. \cite{nemirovsky2023implementation} use resting-state fMRI to estimate $\Phi$ across brain networks, finding higher integration in conscious states compared to unconscious ones, such as during anesthesia. Both studies provide empirical backing for IIT's predictions in humans, leveraging direct neural recordings and imaging to link $\Phi$ with consciousness, though they highlight the challenge of scaling these methods to whole-brain analysis.

Recent literature has increasingly applied IIT to investigate whether LLMs — such as GPT-4 or LLaMA — might be conscious, given their complex architectures and human-like outputs. In general, IIT posits that software-based AI systems cannot achieve consciousness, as they lack the intrinsic cause-effect power required to integrate information effectively \citep{tononi2008consciousness, tononi2015consciousness, tononi2016integrated, koch2019feeling}. \cite{gams2024evaluating} assess ChatGPT's consciousness using IIT alongside its performance in a Turing Test framework. They argue that despite ChatGPT's ability to mimic human-like conversation — potentially passing a behavioral test — its transformer-based architecture lacks the recurrent, integrated causality required for high $\Phi$ under IIT. Their analysis concludes that ChatGPT's low integration precludes consciousness, aligning with IIT's structural criteria, though they note its linguistic proficiency challenges purely functional definitions of awareness. This study underscores IIT's role in distinguishing behavioral competence from subjective experience in LLMs. A recent work from \cite{findlay2024dissociating} applies IIT to distinguish between intelligence and consciousness in artificial systems, with a focus on LLMs. They argue that while LLMs exhibit impressive linguistic competence, their feedforward architectures and lack of intrinsic causal integration yield low $\Phi$, rendering them unconscious under IIT. The study emphasizes the need for bidirectional, reentrant processing — absent in current LLMs — to achieve consciousness, proposing that IIT can serve as a diagnostic tool to dissociate functional AI capabilities from subjective experience. This work reinforces skepticism about LLM consciousness while highlighting IIT's utility in clarifying the intelligence-consciousness divide.

However, most research applying IIT \citep{tononi2015consciousness, koch2019feeling, gams2024evaluating, findlay2024dissociating} has focused primarily on the architecture, structure, topology, or processing mechanisms of LLMs. Existing studies are largely theoretical and lack sufficient quantitative evidence derived from real-world LLMs. Meanwhile, a widely held perspective — independent of IIT — suggests that contemporary Transformer-based LLMs cannot possess consciousness if assessed solely based on their architecture, structure, topology, or processing mechanisms. This view is supported by two key observations: 1) once an LLM completes pre-training and fine-tuning, its parameters remain fixed, and 2) LLMs operate under a mechanistic "next-token prediction" paradigm governed by probabilistic calculations. The paper from \cite{bachmann2024pitfalls} highlights how the fixed computational steps in next-token prediction limit the model's ability to engage in tasks requiring iterative reasoning, implying a lack of the flexible, self-aware processing often linked to consciousness. The survey from \cite{minaee2024largelanguagemodelssurvey} frames LLMs as tools for pattern recognition and generation, not as systems capable of subjective experience, aligning with the view that fixed parameters and probabilistic token prediction preclude consciousness. The focus on LLMs as fixed-parameter, probability-driven predictors highlights their lack of the adaptive, introspective qualities typically linked to consciousness, as proposed by \cite{shlegeris2024language}. There is also a post on X from \cite{lecunyann2024X}, stating that "The problem isn't that it is a transformer. The problem is that it is an auto-regressive LLM. Auto-regressive LLMs that compute each token with a fixed number of computational steps can't reason, regardless of the details of the architecture." And this statement ties the fixed-step, probabilistic prediction process to a broader limitation, supporting the argument that consciousness cannot arise in such systems.

Given the black-box nature of deep learning and the limited understanding of LLM representations, this study seeks to address this knowledge gap by investigating potential "consciousness" phenomena within LLM representations.

In this study, we hypothesize that a network emerges from the LLM representation, where each dimension corresponds to a node within the network, and latent interconnections (edges) exist between individual nodes or clusters of nodes. When analyzed as a sequence (time series) of representations, each time point (representation) represents a distinct state or snapshot of this evolving network. We term this network as the \textbf{Representation Network (RN)}.

For example, a response to a ToM task is expressed in natural language. When this response is fed into an LLM, it generates a sequence of representations\footnote{Representations derived from responses alone are insufficient for observing "consciousness." A performance rating for a response is meaningless without considering the ToM test question (stimulus) it aims to address. Therefore, both the response and its context (i.e., the corresponding stimulus) were considered, as detailed in Sec. \ref{Materials and Methods Overview}, Sec. \ref{Methods_IIT_for_LLM_Representations_Time_series_Signals_from_Attended_Response_Representations}, and Sec. \ref{Methods_IIT_for_LLM_Representations_Response_Representations_Attended_to_Stimulus_Linguistic_Spans_in_Context}.}. This sequence of representations forms a time series, as each individual representation (time point) follows the order of tokens in the original response. By applying the settings for nodes and edges, this sequence of representations gives rise to an RN.

This RN represents an observed, contextualized (potential) "experience" corresponding to the response and its stimulus. Each RN is uniquely determined by the sequence but is independent of the LLM's architecture, structure, or topology once the representation is generated. Consequently, the LLM itself cannot "experience" what it generates and, therefore, cannot possess consciousness. However, a higher-level system that produces or processes representations beyond the "next-token prediction" paradigm — such as an agentic AI system built on top of LLMs — could potentially be capable of "experiencing" these representations, reinforcing the significance of our primary research question.

For each individual RN derived from an LLM's response to a ToM task, we formulated the five IIT axioms applicable to the "subject's" ToM-related "brain" network (RN), assuming that the phenomenon of "consciousness" can be observed from the corresponding sequence of representations.

\begin{itemize}
    \item \textbf{Existence}: When "subjects" perform ToM tasks, their "brain" activity (captured via RN) reflects a conscious state that exists for them — i.e., the "subjective experience" of inferring another's mental state.
    \item \textbf{Composition}: ToM performance involves composing multiple mental elements (i.e., self-perspective, other's perspective, context) into a coherent experience, reflected in distributed "brain" activity.
    \item \textbf{Information}: Each ToM task requires the "brain" to generate a specific, informative conscious state, distinguishing, say, a false belief from a true one.
    \item \textbf{Integration}: ToM requires integrating self-other perspectives, emotions, and context into a unified mentalizing "experience", directly tied to $\Phi$.
    \item \textbf{Exclusion}: During ToM tasks, the "brain" settles into a single, maximally integrated state specific to the task, excluding alternative states.
\end{itemize}

The central question to address our primary one is: \textit{How can we determine whether a given sequence of representations is encoded with an "experience" or a phenomenon of "consciousness"?} While IIT provides a compelling framework for understanding the conscious substrate of ToM, it does not offer a comprehensive explanation. Specifically, IIT accounts for the structural and capacity-related aspects of consciousness but does not fully elucidate the cognitive mechanisms underlying ToM. Consequently, it serves as a complementary rather than a standalone theory of mentalizing. Moreover, IIT alone cannot serve as a definitive criterion, as its panpsychist implications challenge its specificity \citep{chalmers2017combination}. Given these limitations, we adopt a hybrid approach in our study.

Furthermore, the identification of a "consciousness" phenomenon cannot be established through a single case analysis within an RN. Instead, conclusions must be drawn from large-scale, repeated experiments spanning multiple Transformer layers across diverse large language models (LLMs) and incorporating a variety of ToM tasks. To address our primary research question systematically, we employ a triangulated approach, decomposing it into three key criteria for identifying potential manifestations of "consciousness" observed from the sequence of representations generated by LLMs.

\begin{itemize}
    \item \textbf{Criterion 1}: Can estimates of $\Phi$, the primary metric of IIT, robustly differentiate responses across distinct ToM performance levels?
    \item \textbf{Criterion 2}: Do these distinctions remain robust across diverse ToM stimuli in repeated large-scale trials?
    \item \textbf{Criterion 3}: Can IIT estimates provide a stronger basis for interpreting variations in ToM performance than the intrinsic characterization of the sequence of representations, which is independent of any consciousness estimate?
\end{itemize}

A crucial \textbf{caution} must be noted: even if a case successfully meets all three criteria outlined above, this does not necessarily indicate that the corresponding sequence of representations is conscious. Rather, it suggests the observation of a \textit{potential} "consciousness" phenomenon within these representations — nothing more. The RN itself remains a hypothetical construct; it neither "experiences" the world nor functions as the network of any real-world system or subject, at least given the current ways in which we interact with LLMs. However, we remain open to future scenarios in which agentic AI systems, built upon LLMs, produce and consume representations in a manner other than "next-token prediction".

Our goal is to advance the understanding of representational learning paradigms in contemporary Transformer-based LLMs through the lens of consciousness studies. Methodologically, our proposed hybrid approach — integrating ToM, IIT, and interpretation methods for LLM representations — marks the inauguration of this exploration.

\section{Materials and Methods}\label{Materials and Methods}

\subsection{Overview of Materials and Methods}\label{Materials and Methods Overview}

In line with recent advancements, this study focuses on IIT (Version 3.0 and 4.0) and its principal metric, integrated information ($\Phi$) \citep{oizumi2014phenomenology, albantakis2023integrated}

\begin{figure}[ht!]
\centering
\includegraphics[width=1.0\columnwidth, trim={0 3.0cm 0 0},clip]{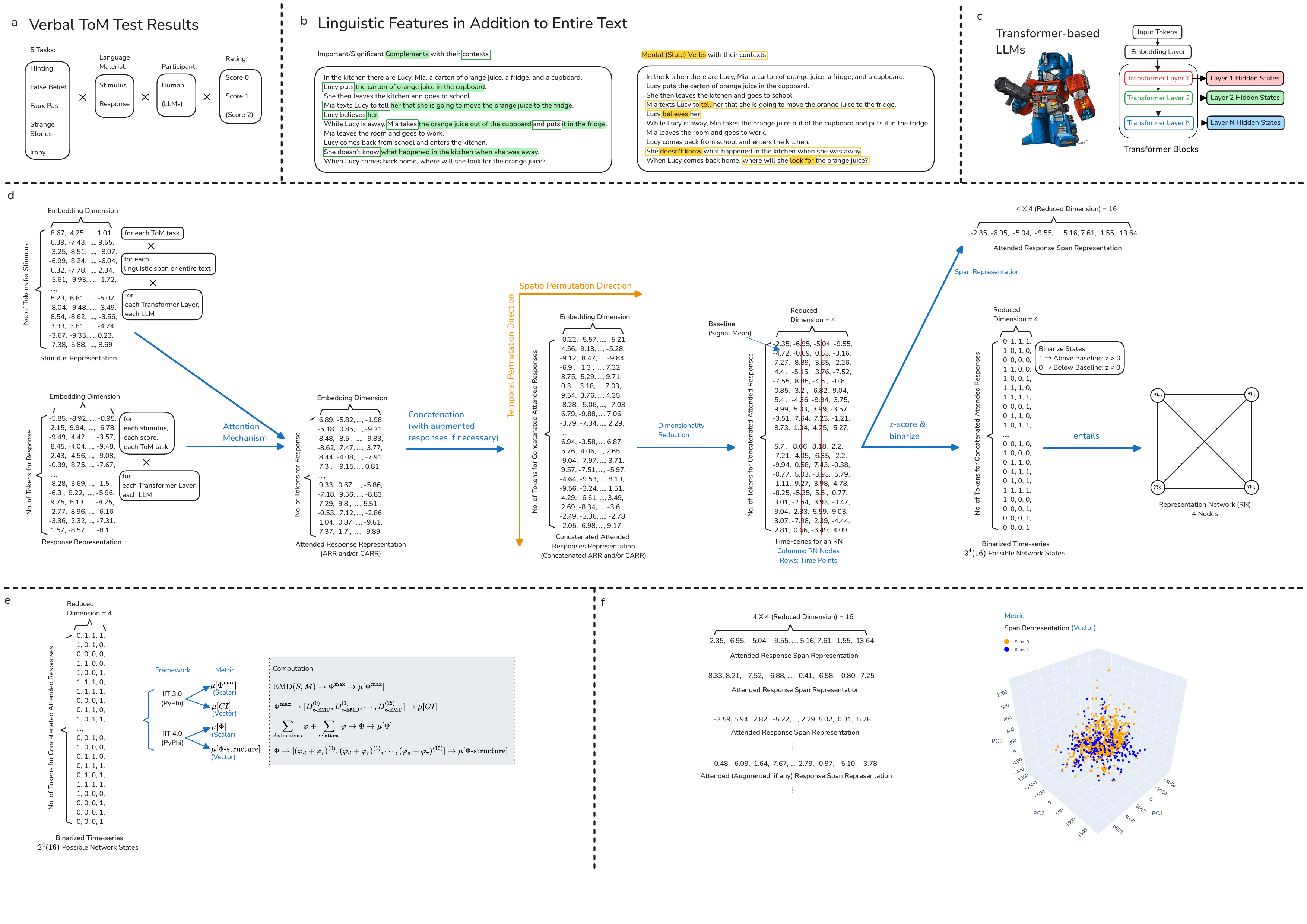}
\caption{\textbf{Summary of Acquisition, Signal Processing, and Metric Extraction.} \textbf{(a)} The ToM test results were sourced from \cite{strachan2024testing}, and the associated dataset is available at \url{https://osf.io/dbn92}. The ToM test comprises five tasks, with each response rated on a scale of 0/1 or 0/1/2. Two types of participants were included in the test: humans and LLMs. We focused exclusively on human responses as discussed in Sec. \ref{Materials and Methods Overview}. We further analyzed these human responses using our selected LLMs: LLaMA3.1-8B, LLaMA3.1-70B, Mistral-7B, and Mixtral-8x7B, the most recent models available at the time of this study. \textbf{(b)} We utilized the web clients of GPT-4o, Claude 3.5 Sonnet, and Gemini to assist in extracting complement syntax and/or mental (state) verbs, where applicable, and labeled the significant ones along with their contexts for each stimulus from the respective ToM tasks. \textbf{(c)} All of the LLMs examined in this study are built using Transformer blocks as their core components. We analyzed multiple layers, not just the final layer, for each model. \textbf{(d)} We obtained the (C)ARR for \textbf{each score category, for each stimulus in every ToM task, spanning both the entire text and specific linguistic spans, as well as across all transformer layers of each LLM}. Dimensionality reduction was then applied to each representation using Principal Component Analysis (PCA), reducing the dimensionality to four (nodes) per time series. We proposed that the RN is embedded within the time series. Each node's time series was binarized with respect to its own mean signal, enabling each RN to assume 16 or $2^4$ possible states at any given time point. To extend the signal length, we concatenated the time series from individual (C)ARR when necessary. Additionally, we derived span representations from the (C)ARR, following the methods outlined in \cite{peters-etal-2018-dissecting} and \cite{jawahar-etal-2019-bert}. These span representations also served as vector metrics, contrasting with the quantities used to estimate consciousness using IIT, as illustrated in \textbf{(f)}. \textbf{(e)} Following the approach outlined in \cite{nemirovsky2023implementation}, we computed the scalar metrics $\mu[\Phi^{\max}]$ (IIT 3.0) and $\mu[\Phi]$ (IIT 4.0) for each RN. Additionally, for each state of the RN, we calculated $CI$ (IIT 3.0) by decomposing $\Phi^{\max}$ (IIT 3.0) into a vector of extended earth mover's distance ($D_{\text{e-EMD}}$), where each component corresponded to a state treated as the mechanism. Similarly, $\Phi$ (IIT 4.0) was decomposed into a vector of distinctions and relations, $\Phi{\text{-structure}} = \varphi_d + \varphi_r$, each associated with their corresponding RN state, when treated as the mechanism. Finally, we computed the two vector metrics for each state and derived the weighted average over the time series.}\label{methods_overview}
\end{figure}

The LLM representations were derived using language data obtained from the ToM test results \citep{strachan2024testing} (Fig. \ref{methods_overview}a). The corresponding dataset is publicly accessible at \url{https://osf.io/dbn92}. We analyzed data from five distinct ToM tasks: the Hinting Task, the False Belief Task, the Recognition of Faux Pas\footnote{We were unable to process the Faux Pas task due to the limitation of the data provided by the original study. Consequently, this task was excluded from our investigation. Further details can be found in Appendix A.2.3.}, Strange Stories, and Irony Comprehension. The dataset is organized by task, with each sheet containing multiple questions (stimuli) and corresponding responses from participants, including humans and large language models (LLMs) such as GPT-4, GPT-3.5, LLaMA2-70B, LLaMA2-13B, and LLaMA2-7B. We focused exclusively on the human responses in the dataset and further investigated their representations using our selected LLMs, since:

\begin{itemize}
    \item The LLMs from the original study and released dataset fall outside the scope of the current study.
    \item We did not investigate how responses are generated by the LLM, as this would pertain to the study of its language generation mechanism — a line of inquiry unlikely to yield meaningful findings, given that consciousness cannot emerge from the "next-token prediction" paradigm. Instead, our focus was on analyzing the sequences of representations formed when linguistic inputs are presented to the LLM.
    \item Most importantly, the linguistic materials used to derive LLM representations must originate from real human subjects who are confirmed to be conscious during testing. Specifically, these subjects are adult native English speakers (aged 18–70) with no history of psychiatric conditions or dyslexia, as specified in \cite{strachan2024testing}. This ensures that variations in ToM test performance (classified as good, medium, or poor) and corresponding differences in IIT estimates ($\Phi$) within the ToM-related "brain" network can be meaningfully interpreted.
\end{itemize}

Our selected models include LLaMA3.1-8B, LLaMA3.1-70B, Mistral-7B, and Mixtral-8x7B, which are the most recent LLMs at the time of this study. All selected models are open source and hosted on Hugging Face (\url{https://huggingface.co/}). For further details, please refer to Appendix A.1.

In our study, an additional set of LLMs was employed to assist with data processing, specifically for labeling different linguistic features in stimuli (Sec. \ref{Methods_IIT_for_LLM_Representations_Response_Representations_Attended_to_Stimulus_Linguistic_Spans_in_Context}) and text augmentation for responses (Sec. \ref{Methods_IIT_for_LLM_Representations_Text_Augmentation_for_Responses}). We interacted with these models via the web interfaces of GPT-4o\footnote{\url{https://chatgpt.com/}, using the version gpt-4o-2024-08-06}, Claude 3.5 Sonnet\footnote{\url{https://claude.ai/}, using the version claude-3-5-sonnet@20240620}, and Gemini\footnote{\url{https://gemini.google.com/}, using the version google/gemini-1.5-flash-002}. However, due to their closed-source nature, we were unable to obtain their internal representations, and therefore, they were excluded from our investigation.

However, the ToM test results were unevenly distributed across scores (performances) for each stimulus within each ToM task. The scores are categorized with 0/1/2 for the Strange Stories task and 0/1 for other tasks. We included only those stimuli where responses contained at least two distinct scores: both Score 0 and Score 1, or Scores 1 and 2 for the Strange Stories task. Responses consisting of single words, such as "Yes" or "No," were excluded due to their limited representational variation and significance when treated as time series data, and therefore were not considered after the stimuli filtering process. For further details, please refer to Appendix A.2.

We propose that a \textbf{Representation Network (RN)} is inherent in the \textbf{Attended Response Representations (ARR)} and the \textbf{Contextually Attended Response Representations (CARR)} when treated as time series. This RN serves as the basis for estimating integrated information ($\Phi$). To align with the discrete element requirements of IIT 3.0 and 4.0, each RN's time series was standardized and binarized. And we established node-specific thresholds based on individual mean signal strengths (Fig. \ref{methods_overview}c) inspired by \cite{nemirovsky2023implementation}. We utilized PyPhi software \citep{oizumi2014phenomenology, mayner2018pyphi, albantakis2023integrated} to compute $\Phi^{\max}$ (IIT 3.0) and $\Phi$ (IIT 4.0). Given that these estimates are state-dependent and temporally variable, we calculated their weighted averages across each network's time series, denoted as $\mu[\Phi^{\max}]$ (IIT 3.0) and $\mu[\Phi]$ (IIT 4.0).

\footnote{See Appendix A.3 and Appendix A.4 for brief introductions to neuroimaging techniques, i.e., fMRI and EEG, and brain cortical regions, i.e., mPFC, TPJ, PCC, and STS.}When applied to different ToM tasks, IIT would attribute differences in ToM test performance (good, medium, poor) to variations in $\Phi$ — the degree of information integration — within and across ToM-related brain networks. Neuroimaging data during these tasks would reveal these differences as variations in activation, connectivity, and synchrony, which IIT interprets as proxies for integration. The followings are grounded in empirical studies linking specific ToM tasks to neuroimaging findings (fMRI activation/connectivity, EEG synchrony), which align with IIT's emphasis on integration ($\Phi$).

Good ToM performance is predicted with high $\Phi$ according to IIT, reflecting strong integration across mPFC, TPJ, STS, and PCC, enabling complex mental state attribution, i.e., 

\begin{itemize}
    \item Hinting Task: Successful hint interpretation requires integrating context, intent, and social norms. fMRI might show robust mPFC-TPJ connectivity \citep{frith2006neural}, and EEG could reveal high gamma-band synchrony, indicating integrated processing \citep{dumas2010inter}.
    \item False Belief Task: Passing (i.e., understanding Sally's false belief) demands self-other differentiation. fMRI shows TPJ and mPFC activation \citep{saxe2003people}, with high connectivity suggesting elevated $\Phi$ \citep{saxe2004understanding}.
    \item Recognition of Faux Pas: Detecting a faux pas integrates emotional and cognitive cues. fMRI might reveal strong PCC-TPJ-STS coupling \citep{stone1998frontal}.
    \item Strange Stories: Advanced mentalizing (i.e., interpreting a lie) requires recursive reasoning. fMRI shows widespread ToM network activation \citep{griffin2006theory}.
    \item Irony Comprehension: Grasping sarcasm integrates intent and tone. fMRI indicates STS and mPFC engagement \citep{uchiyama2006neural}, with EEG showing cross-frequency coupling (i.e., theta-gamma) \citep{canolty2006high}.
\end{itemize}

Medium ToM performance is predicted with moderate $\Phi$ according to IIT, which is sufficient for basic mentalizing but not complex recursion or nuance, due to partial integration in the ToM network. i.e., 

\begin{itemize}
    \item Hinting Task: Partial success (i.e., literal interpretation) might show mPFC activation but weak TPJ coupling on fMRI \citep{corcoran1995schizophrenia}.
    \item False Belief Task: Failure (i.e., in 3-year-olds) reflects incomplete self-other integration. fMRI might show TPJ activation without strong mPFC connectivity \citep{sabbagh2009neurodevelopmental}, and EEG could lack sustained coherence \citep{liu2009neural}.
    \item Recognition of Faux Pas: Missing subtle blunders suggests emotional-cognitive disconnect. fMRI might reveal STS activity but poor PCC integration \citep{shamay2005neuroanatomical}, with EEG showing fragmented theta \citep{nigbur2012theta}.
    \item Strange Stories: Basic comprehension without deeper insight could show moderate mPFC-STS activation on fMRI \citep{fletcher1995other}.
    \item Irony Comprehension: Literal interpretation might involve STS but not mPFC integration \citep{wang2006neural}, with EEG showing weak cross-frequency coupling \citep{canolty2006high}.
\end{itemize}

Poor ToM performance will result in low or disrupted $\Phi$ in the ToM network according IIT, impairing the ability to generate distinct mental state representations, i.e., 

\begin{itemize}
    \item Hinting Task: Failure to infer intent might show isolated mPFC or STS activation on fMRI \citep{muller2018brain}, with EEG lacking synchrony, indicating low $\Phi$ \citep{just2004cortical}.
    \item False Belief Task: Failure (i.e., in severe autism) could reflect minimal TPJ-mPFC integration on fMRI \citep{kana2009atypical}, with EEG showing desynchronized patterns \citep{murias2007resting}.
    \item Recognition of Faux Pas: Missing faux pas entirely might involve weak PCC-STS connectivity on fMRI \citep{shamay2005neuroanatomical}, with EEG lacking theta/gamma coherence \citep{just2004cortical}.
    \item Strange Stories: Inability to mentalize could show sparse ToM network activation on fMRI \citep{winner1998distinguishing}, with EEG indicating low integration (i.e., delta dominance) \citep{murias2007resting}.
    \item Irony Comprehension: Literal or absent understanding might involve STS activation without mPFC coupling on fMRI \citep{wang2006neural}, with EEG showing no cross-frequency integration \citep{murias2007resting}.
\end{itemize}

It is important to emphasize that the absolute or isolated value of any IIT estimate lacks intrinsic meaning. Such estimates are only interpretable when compared across varying levels of consciousness (i.e., degree) and/or different conscious contents (i.e., qualia), assuming the comparisons are conducted under consistent contextual conditions \citep{oizumi2014phenomenology, albantakis2023integrated}. For example, a high value of an IIT metric — whether $\Phi^{\max}$ (IIT 3.0) or $\Phi$ (IIT 4.0) — does not, in itself, indicate that the underlying network is conscious. In the study applying IIT to human resting-state fMRI data \citep{nemirovsky2023implementation}, the analysis and findings were derived from IIT estimates compared across four awareness conditions — Awake, Mild sleep, Deep sleep, and Recovery (from sleep) — for each individual subject under their specific health condition.

When applied to our research context, IIT estimates are meaningful only when compared across different ToM test performances, provided that all other variables are held constant — such as the Transformer layer sampled, the LLM used, the linguistic span, the ToM task, and the permutation control etc.

To address the primary research question, we investigated whether the estimates for $\mu[\Phi^{\max}]$ (IIT 3.0) and $\mu[\Phi]$ (IIT 4.0) could reliably differentiate between performances of the ToM test, across varying score ratings assigned to each response, as discussed above (see Sec. \ref{Methods_IIT_for_LLM_Representations_Measuring_Integrated_Information} for details). Furthermore, we extended our analysis to identify which transformer layers of the LLMs (Fig. \ref{methods_overview}c), as well as which linguistic spans and their contexts (Fig. \ref{methods_overview}b), exhibited significant variations in these estimates when attending to the stimuli. If LLM representations are observed with "consciousness" phenomenon, they should robustly discriminate between score categories, with higher scores reflecting enhanced activations and/or connectivity in ToM-related "brain" networks. Consequently, higher ratings would be expected to be observed with elevated values of $\mu[\Phi^{\max}]$ (IIT 3.0) and/or $\mu[\Phi]$ (IIT 4.0), as predicted by IIT \citep{oizumi2014phenomenology, albantakis2023integrated}.

As this study explores the implementation of IIT 3.0 and IIT 4.0, we compared $\mu[\Phi^{\max}]$ (IIT 3.0) and $\mu[\Phi]$ (IIT 4.0) with other metrics potentially interpreting the performance differences revealed in ToM test results. We introduced two additional metrics (Fig. \ref{methods_overview}e) in our analysis (see Sec. \ref{Methods_IIT_for_LLM_Representations_Measuring_Integrated_Information} for details): 1) the Conceptual Information ($CI$) of the specific partitioned constellation from which $\Phi^{\max}$ is determined in IIT 3.0 \citep{oizumi2014phenomenology}, and 2) the $\Phi$-structure, defined as the cause–effect structure specified by a maximal substrate in IIT 4.0 \citep{albantakis2023integrated}. These metrics decompose the scalar values of $\Phi^{\max}$ (IIT 3.0) and $\Phi$ (IIT 4.0), respectively, into vectors associated with each state when treated as the mechanism in the RN. 

Since IIT alone is insufficient to give a complete explanation to the differences in ToM test performances as discussed in Sec. \ref{Introduction}, we further characterized the ToM test results with additional quantities, such as the Span Representation (Fig. \ref{methods_overview}f), derived from the same (dimensionality reduced) LLM representations, used to estimate consciousness. This additional effort aims to act as the third pillar interpreting the ToM test results by differentiating between observed potential "consciousness" phenomena and (well) separated manifolds inherent in the LLM representations. The two additional vector-based metrics introduced in our study — Conceptual Information ($CI$) from IIT 3.0 and $\Phi$-structure from IIT 4.0 — are of particular importance, as they provide more informative and distinguishable insights into differences in ToM test performance compared to a single scalar IIT estimate, such as $\Phi^{\max}$ (IIT 3.0) or $\Phi$ (IIT 4.0). In this regard, they complement and/or contrast with comparisons between Span Representation and scalar-based IIT estimates.

\subsection{Deriving Estimates of Consciousness from An Implementation of Integrated Information Theory for LLM Representations}\label{Methods_IIT_for_LLM_Representations}

The \textbf{Representation Network (RN)} inherent in the sequence of LLM representations by LLMs serves as the basis to derive IIT estimates. The RN is a graph with $D$ nodes, and the sequence of representations has the shape $(n, D)$, where $n$ is the number of tokens in the sequence and $D$ is the embedding dimension for a particular LLM.

\subsubsection{Time series Signals from Attended Response Representations (ARR)}\label{Methods_IIT_for_LLM_Representations_Time_series_Signals_from_Attended_Response_Representations}

The representations derived solely from responses are insufficient to serve as a basis for observing "consciousness." The performance rating of a response is meaningless without reference to the specific ToM test question (stimulus) it attempts to answer. Only when a response is analyzed within its context — specifically, the corresponding ToM test question — do the associated representations acquire meaningful significance.

Both the representations for the stimuli and responses are time series. Using these time series, we constructed the Transition Probability Matrix (TPM) (see Sec. \ref{Methods_IIT_for_LLM_Representations_Constructing_the_Transition_Probability_Matrix}) and estimated the Integrated Information for the time series of each response, given that of its corresponding stimulus (see Sec. \ref{Methods_IIT_for_LLM_Representations_Measuring_Integrated_Information}).

For each of the selected LLMs, we fed the stimuli and their corresponding responses into the LLM to obtain the representations. For example, for a particular stimulus, the tokenization by the LLM's tokenizer resulted in $n_1$ tokens, with each token having $D$ dimensions, representing the embedding dimension of the LLM. Correspondingly, the $k$-th response may have $n_{2,k}$ tokens, each also with the same $D$ dimensions.

In order to put the response into its context, i.e., the specific ToM test question (stimulus) it attempts to answer, we applied attention mechanisms \citep{NIPS2017_3f5ee243} to associate each response with its corresponding stimulus in the following steps:

1. \textbf{Dot Product Attention}: This is the form of attention we employed where the response tokens "attend" to the stimulus tokens. This attention mechanism works by computing a score that measures the relevance of each stimulus token to each response token.

2. \textbf{Calculating Attention Scores}:

\begin{equation}
S_{\text{Attention}} = \frac{R_{\text{Response}} \cdot R_{\text{Stimulus}}^T}{\sqrt{D}} \label{Attention_Scores}
\end{equation}

where,

\begin{align}
S_{\text{Attention}} \text{ is the Attention Scores, and is of shape } (n_{2,k}, n_1); \nonumber \\
R_{\text{Response}} \text{ is the Response Representation, and is of shape } (n_{2,k}, D); \nonumber \\
R_{\text{Stimulus}} \text{ is the Stimulus Representation, and is of shape } (n_1, D); \nonumber
\end{align}

The attention scores for each pair of response and stimulus tokens are calculated as a dot product between the corresponding token representations, followed by scaling by $\sqrt{D}$ to normalize the scores.

The resulting Attention Scores matrix has a shape of $(n_{2,k}, n_1)$. This means that for each of the $n_{2,k}$ tokens in the response, The set of $n_1$ scores indicating how much attention should be paid to each of the $n_1$ tokens in the stimulus.

3. \textbf{Applying Softmax to Attention Scores}: After computing the raw attention scores, a softmax function is applied to convert these scores into probabilities. The softmax is applied along the dimension corresponding to the stimulus tokens:

\begin{equation}
W_{\text{Attention}} = \text{softmax}(S_{\text{Attention}}) \label{Attention_Weights}
\end{equation}

where,

\begin{align}
W_{\text{Attention}} \text{ is the Attention Weight, and is of shape } (n_{2,k}, n_1); \nonumber \\
S_{\text{Attention}} \text{ is the Attention Scores from Eq. \ref{Attention_Scores}, and is of shape } (n_{2,k}, n_1); \nonumber
\end{align}

The Attention Weights retain the shape $(n_{2,k}, n_1)$, where each row (corresponding to a response token) sums to $1$.

4. \textbf{Computing Attended Response Representation (ARR)}: The final attended representation for each response token is obtained by taking a weighted sum of all stimulus token representations, weighted by the attention scores:

\begin{equation}
R_{\text{Attended Response},i} = \sum_{j=1}^{n1} W_{\text{Attention},i,j} \cdot R_{\text{Stimulus},j} \label{Attended_Response_Representation}
\end{equation}

where,

\begin{align}
R_{\text{Attended Response},i} \text{ is the } i\text{-th Attended Response Representation;} \nonumber \\
W_{\text{Attention},i,j} \text{ is the Attention Weight between }i \text{-th response and } j\text{-th stimulus;} \nonumber \\
R_{\text{Stimulus},j} \text{ is the } j\text{-th Stimulus Representation;} \nonumber
\end{align}

This operation is performed for each response token, resulting in an attended representation.

The resulting \textbf{Attended Response Representation (ARR)} will have a shape of $(n_{2,k}, D)$. This shape is the same as the original response representation, but now each token in the response has been modified based on the context provided by the stimulus.

Each ARR corresponds to \textbf{a specific score category for a given stimulus from a ToM task, across each of the transformer layers of each of the LLMs} (See Sec. \ref{Methods_IIT_for_LLM_Representations_Representations_from_Multiple_Transformer_Layers} for details).

\textbf{Intuition Behind the Attention Mechanism}. The attention mechanism allows the response to dynamically focus on different parts of the stimulus when analyzing each token. This focus can vary depending on the specific token in the response, leading to a more contextually informed and potentially more accurate representation. This is especially useful in our study, or any scenario where understanding the relationship between different sequences (like a stimulus and a response) is crucial.

In addition to Attention Mechanism, we in fact came up with several alternative approaches, i.e., 

\begin{itemize}
    \item \textbf{Concatenation of Representations}: Concatenate the stimulus and response representations along the sequence dimension to create a single combined representation.
    \item \textbf{Mean Pooling and Vector Operations}: Compute the mean representation of the stimulus and the response separately and then perform operations like addition, subtraction, or dot product on the mean vectors.
    \item \textbf{Residual Connection}: Add the stimulus' representation to the response's representation directly, using a residual connection to maintain the stimulus' influence in the response.
    \item \textbf{Joint Embedding Space (Pretrained)}: Use a model that has been pretrained to embed stimuli and responses into a shared space.
\end{itemize}

The Attention Mechanism remains the best fit, this is because:

\begin{enumerate}
    \item \textbf{Attention Preserves Temporal and Dimensional Structure}
    \begin{itemize}
        \item Attention computes fine-grained, token-wise influence: how each response token attends to specific stimulus tokens.
        \item This preserves the temporal nature of the response — we don't collapse time, and we preserve per-token RN state evolution.
        \item We get a dynamic, interpretable mapping between the dimensions (nodes) in the response representation under the stimulus context.
    \end{itemize}
    \item \textbf{Attention is Context-Adaptive}
    \begin{itemize}
        \item Unlike static methods (i.e., concatenation or averaging), attention is context-sensitive: each response token can draw from different stimulus tokens depending on meaning and position.
        \item This emergent structure is critical for modeling RN as a dynamic and evolving network. We captured token-specific influence patterns, reflecting emergent topology in the RN.
    \end{itemize}
\end{enumerate}

Each alternative approach presents specific limitations, i.e.,

\begin{enumerate}
    \item \textbf{Concatenation of Representations}
    \begin{itemize}
        \item Treats the stimulus and response as flat, joined sequences.
        \item No mechanism to highlight or control influence of stimulus over response.
        \item Loses token-specific interaction and introduces artificial temporal flattening.
        \item Not suitable for modeling directional influence or latent interactions in RN.
    \end{itemize}

    \item \textbf{Mean Pooling or Vector Operations}
    \begin{itemize}
        \item Reduce the stimulus to a single vector.
        \item Every response token sees the same summary, ignoring token-specific semantics.
        \item Destroys the time-resolved interaction critical to representation analysis.
        \item Collapses structural information, which is incompatible with node-edge evolution modeling.
    \end{itemize}

    \item \textbf{Residual Connections}
    \begin{itemize}
        \item Simple addition of stimulus and response vectors.
        \item No distinguishable attention or selective focus — treats all stimulus info as equally important.
        \item No control over which dimensions (nodes) are influenced or how.
        \item Oversimplifies influence, which means it does not support interpretable edge structure in RN.
    \end{itemize}

    \item \textbf{Joint Embedding Space (Pretrained)}
    \begin{itemize}
        \item Assumes global compatibility, but doesn't model token-to-token influence.
        \item We lose sequential dependencies and per-token RN dynamics.
        \item The introduction of additional artifacts may compromise the primary objective of interpretation — namely, the learned representations — by shifting focus toward artifacts generated by the pretrained model.
    \end{itemize}
\end{enumerate}

The Attention Mechanism remains the most faithful, structured, and theoretically grounded method for incorporating stimulus influence into the response's representation leading to an RN. It models temporal, token-level, and latent edge interactions, exactly in line with our hypothesis of the RN as a dynamic, evolving latent graph over representation dimensions.

There are several forms of attention mechanisms in addition to the dot-product attention adopted in our approach (a.k.a. scaled dot-product attention, as introduced in \cite{NIPS2017_3f5ee243}. However, even after considering these alternatives, dot-product attention remains the best fit for our study context — where each representation dimension is a node, and our RN emerges from a time series of high-dimensional token representations.

The primary reason for excluding other forms of attention mechanisms is that introducing a learned function to embed stimuli and responses into a shared space would risk generating additional artifacts. These artifacts could obscure the true target of interpretation — mirroring the limitations presented in the \textit{Joint Embedding Space (Pretrained)} approach discussed above.

For each specific alternative, i.e.,

\begin{itemize}
    \item The \textbf{Additive Attention} introduced in \cite{bahdanau2014neural} is computationally more expensive (matrix operation + nonlinearity) and less parallelizable. And the learned function complicates interpretability.
    \item The \textbf{Cosine Attention} \citep{graves2014neural}, using cosine similarity as the attention score between $q$ and $k$, removes the magnitude info, which could be meaningful in transformer embeddings. Our RN treats each dimension as a node indicating magnitude may carry useful node-specific signal.
    \item The \textbf{Multi-head Attention (MHA)} \citep{NIPS2017_3f5ee243} is an extension, not a replacement of dot-product attention. However, We could not determine the optimal number of heads prior to implementing MHA.
    \item The \textbf{Learned Attention (i.e., Linear Attention \citep{katharopoulos2020transformers}, Attention with Gating \citep{xue2020not}, etc.)} is designed for \textit{performance}, not interpretability. And they obfuscate the explicit interaction structure between token representations that we modeled in RN.
\end{itemize}

Although there are other attention variants (additive, cosine, multi-head, gated), the dot-product attention mechanism remains the best-suited and most interpretable choice for our proposed RN because:

\begin{itemize}
    \item It supports dynamic, per-token influence.
    \item It provides a natural and interpretable similarity measure between high-dimensional vectors.
    \item It preserves the fine-grained, temporal structure needed to study the RN as an evolving network over time.
\end{itemize}

Considering the scope and depth of our study, we will optionally extend our analysis in future to \textit{multi-head attention} for broader patterns — but dot-product attention is the most principled starting point.

\subsubsection{Response Representations Attended to Stimulus Linguistic Spans in Context (CARR)}\label{Methods_IIT_for_LLM_Representations_Response_Representations_Attended_to_Stimulus_Linguistic_Spans_in_Context}

Previous studies have highlighted the influence of linguistic features on ToM tests \citep{astington1999longitudinal, de2002complements, de2007interface, milligan2007language, de2014linguistic}. Accordingly, we examined this aspect alongside the entire spectrum of each stimulus. We utilized the web interfaces of GPT-4o, Claude 3.5 Sonnet, and Gemini to extract complement syntax and mental (state) verbs, where applicable. The output from each LLM was carefully examined, with our manual labeling of each token representing either the complement or the context in which the complement appears. Text was tokenized using the spaCy tool (\url{https://spacy.io/}), assigning each token an index to quantify the spans of complements and their contexts. The specific token spans for each LLM were mapped accordingly considering the different tokenizers from different LLMs under investigation. A similar approach was applied to label mental (state) verbs and their contexts. Further details can be found in Appendix A.5.

The identification of complements and mental (state) verbs, along with their contexts, varies on a case-by-case basis and is stimulus-specific. Currently, no standardized benchmark exists to determine which specific spans of complements/mental (state) verbs and their contexts are sufficiently significant to substantially influence reasoning in mental or cognitive processes, thereby enabling accurate responses to ToM test questions.

To the best of our knowledge, the most effective approach involves leveraging existing LLMs to initially label specific linguistic spans, followed by synthesizing the results through human validation.

Our analysis extended beyond examining the response to the entire stimulus. We specifically focused on the labeled linguistic spans, including complement syntax and mental (state) verbs. Furthermore, we considered the contextual environment of these complements and mental (state) verbs to provide a comprehensive linguistic analysis.

To emphasize specific spans or words, along with their context within the stimulus, our approach prioritizes assigning greater attention to these elements during the attention calculation. This was achieved by directly adjusting the attention weights, by masking or reducing the influence of other tokens, thereby concentrating the response's focus on the targeted spans.

1. \textbf{Masking Non-relevant Tokens and Context Tokens}: After calculating the attention scores as in Eq. \ref{Attention_Scores}, we applied masks with different values that reduced attention to non-relevant tokens and context tokens respectively:

\begin{equation}
S_{\text{Attention}}' = S_{\text{Attention}} \label{Masked_Attention_Scores_Copy}
\end{equation}

\begin{equation}
S_{\text{Attention}}' \leftarrow S'_{\text{Attention } | \text{ } i, j \notin [p,q] \text{ and } i, j \notin [u,v]} \cdot m_{\text{non-relevant}} \label{Masked_Attention_Scores_non_Relevant}
\end{equation}

\begin{equation}
S_{\text{Attention}}' \leftarrow S'_{\text{Attention } | \text{ } i, j \in [u,v]} \cdot m_{\text{context}} \label{Masked_Attention_Scores_Context}
\end{equation}

where,

\begin{align}
S_{\text{Attention}} \text{ is the Attention Scores from Eq. \ref{Attention_Scores}, and is of shape } (n_{2,k}, n_1); \nonumber \\
S_{\text{Attention}}' \text{ is the Masked Attention Scores, and is of shape } (n_{2,k}, n_1); \nonumber \\
p \text{ and } q \text{ delimit the start and end of our interested spans;} \nonumber \\
u \text{ and } v \text{ delimit the start and end of the contexts} \nonumber \\
\text{ in which our interested spans appear;} \nonumber \\
m_{\text{non-relevant}} = 0.2 \text{ is the mask value for non-relevant tokens;} \nonumber \\
m_{\text{context}} = 0.6 \text{ is the mask value for context tokens;} \nonumber
\end{align}

This strategy forces the response to focus primarily on our interested spans or words in the stimulus (w/o mask value or $m_{\text{interested}} = 1.0$), ensuring that these words heavily influence the response tokens.

The masking values were carefully selected based on three key criteria: \textbf{distinguishability} between different spans, the \textbf{non-triviality}, and \textbf{non-negativity} of the values, i.e.,

\begin{itemize}
    \item \textbf{distinguishability}: Our selected masking value set (1.0, 0.6, 0.2) provides greater distinguishability among the target spans, their surrounding contexts, and non-relevant segments within the stimulus, compared to alternative configurations such as (1.0, 0.9, 0.8), (1.0, 0.8, 0.6), or (1.0, 0.7, 0.4).
    \item \textbf{non-triviality}: Although our study focuses on evaluating the influence of complements and mental (state) verbs within the stimulus, alongside their contexts, tokens outside these spans cannot be overlooked. These tokens are critical for maintaining the integrity and coherence of the stimulus, precluding the simplistic assignment of a zero masking value (i.e., $m_{\text{context}} = 0.0$).
    \item \textbf{non-negativity}: As each token constitutes an integral and coherent component of the stimulus, assigning a negative masking value to $m_{\text{interested}}$, $m_{\text{context}}$, or $m_{\text{non-relevant}}$ would render the results uninterpretable and devoid of meaningful insight.
\end{itemize}

The \textbf{Contextually Attended Response Representations (CARR)} for each stimulus were obtained concurrently with the ARR, as described in Sec. \ref{Methods_IIT_for_LLM_Representations_Time_series_Signals_from_Attended_Response_Representations}. These representations correspond to \textbf{each score category for every stimulus in each ToM task, across all transformer layers from each LLM} (See Sec. \ref{Methods_IIT_for_LLM_Representations_Representations_from_Multiple_Transformer_Layers} for further details).

\subsubsection{Representations from Multiple Transformer Layers}\label{Methods_IIT_for_LLM_Representations_Representations_from_Multiple_Transformer_Layers}

The LLMs examined in this study are primarily composed of Transformer blocks \citep{NIPS2017_3f5ee243}, with each model comprising a distinct number of Transformer layers. Our analysis encompassed multiple layers, including but not limited to the final layer, for each LLM.

We utilized the \href{https://huggingface.co/}{HuggingFace} library to extract representations from multiple Transformer layers of each large language model (LLM) by setting the parameter \verb|output_hidden_states=True|. This approach aligns with prior studies, such as \cite{caucheteux2023evidence}, which examined representations from multiple Transformer layers across various LLMs and compared these with human brain activity during natural language processing.

According to the \href{https://huggingface.co/docs/transformers/main_classes/output}{HuggingFace official documentation}, when setting \verb|output_hidden_states=True| in a model's forward pass, we're instructing the model to return the hidden states from all layers in the transformer architecture, not just the final output representation. This parameter is directly related to both self-attention and feedforward layers. The hidden states returned include:

\begin{itemize}
    \item The input embeddings (initial hidden state before any transformer layers); and
    \item The output of each transformer layer, which reflects the processed representations after self-attention, feedforward operations, normalization (i.e., LayerNorm), and residual connections.
\end{itemize}

Each transformer block typically consists of:

\begin{itemize}
    \item A self-attention mechanism; and
    \item A feedforward neural network; and
    \item Layer normalization components.
\end{itemize}

When setting \verb|output_hidden_states=True|, we get access to the output representation after each complete block has processed the data, giving us visibility into how the representation evolves through the network.

So more precisely speaking, we investigated the representations from each Transformer \textbf{BLOCK} layer, after the process of self-attention mechanism, feedforward layer, and layer normalization etc.

When \verb|output_hidden_states=True| is set for Mixtral-8x7B, a Mixture-of-Experts (MoE) model \citep{jiang2024mixtral}, the model returns a tuple containing:

\begin{itemize}
    \item The input embeddings (after tokenization and embedding lookup); and
    \item The hidden state from each of the model's layers.
\end{itemize}

For each layer, the hidden state is the output after:

\begin{itemize}
    \item \textbf{Self-attention}: Computes contextual relationships between tokens using sliding window attention and Group Query Attention (GQA).
    \item \textbf{MoE Feedforward Block}: For each token:
    \begin{itemize}
        \item The router computes gating scores to select the top-2 experts; and
        \item The selected experts process the token's representation (post-attention); and 
        \item The outputs of the two experts are weighted by their gating scores and summed to produce the final feedforward output.
    \end{itemize}
    \item \textbf{LayerNorm and Residual Connections}: The feedforward output is normalized and added to the attention output (via residual connection), producing the layer's final hidden state.
\end{itemize}

Thus, the hidden state for layer $i$ reflects the combined effect of self-attention and the sparse MoE feedforward computation, where only a subset of experts contributes to each token's representation.

In short, the Transformer layers we sampled are still the Transformer \textbf{BLOCK} layers, and after the effect of self-attention and/or feedforward pass for each LLM under investigation.

Rather than analyzing every individual layer of each LLM, we systematically sampled $12$ layers, evenly distributed from the first to the last (inclusive), for each model. Additionally, we focused on the intermediate-to-deep layers ($l = \frac{2}{3} n_{layers}$) of each LLM, as suggested by \cite{caucheteux2023evidence}. These layers merit particular attention due to their demonstrated superior predictive capabilities for human brain activity \citep{schrimpf2021neural, caucheteux2022brains, caucheteux2023evidence}. For a detailed description of the LLMs selected for this investigation, please refer to Appendix A.1.

To ensure consistency and fairness in comparison, we fixed the sample size at 12 layers for each LLM under investigation. This strategy is justified by aligning the relative positions of the selected layers across models with differing depths. For example, comparing the 24th layer (i.e., 9/12) of LLaMA3.1–8B with the 24th layer of LLaMA3.1–70B (approximately 4/12) would be misleading due to their differing proportions within the total number of layers. Consequently, only the 12 proportionally aligned layers were included in the sampling process, which also supports the adequacy of this sample size for larger models such as LLaMA3.1–70B.

\subsubsection{Text Augmentation for Responses}\label{Methods_IIT_for_LLM_Representations_Text_Augmentation_for_Responses}

The typical length of a response sequence for a given stimulus is insufficient to constitute a valid time series that satisfies both the conditional independence and Markov property assumptions of IIT. To address this limitation, we adopted an approach inspired by \cite{nemirovsky2023implementation}, concatenating all available response representations for each stimulus. However, this method's efficacy remains constrained by the limited number of response representations, even after concatenation, and is thus applicable to only a small subset of stimuli.

To address the insufficient response length, we implemented a response augmentation strategy, aiming to achieve a minimum of $1,000$ words for the concatenated responses (including original responses) for \textbf{each score category across all stimuli and ToM tasks}. The threshold of $1,000$ words was chosen arbitrarily as a sufficiently large sample size within which to search for the optimal time series satisfying conditional independence and Markov property, as detailed in Sec. \ref{Methods_IIT_for_LLM_Representations_Searching_for_Optimal_Conditional_Independence_and_Markov_Property}. This augmentation process utilized the web clients of GPT-4o, Claude 3.5 Sonnet, and Gemini, generating additional text based on each original response while preserving its intended meaning. We rigorously validated each augmented response to ensure accuracy and relevance. For a comprehensive description of the text augmentation methodology, please refer to Appendix A.6.

Because the linguistic content of both original human responses and their augmented counterparts is independent of the LLM under investigation, we employed more advanced models — such as GPT-4o, Claude 3.5 Sonnet, and Gemini — to generate high-quality augmented texts. Although these augmented responses may be viewed as LLM-generated imitations of human responses, we mitigated this limitation through meticulous prompt design and rigorous human review of each output. Furthermore, we ensured that the source text for augmentation was derived exclusively from genuine human responses.

This approach, however, introduces a methodological limitation akin to that reported by \cite{nemirovsky2023implementation}, stemming from the concatenation of distinct response representations for each stimulus. Consistent with \cite{nemirovsky2023implementation}, we anticipate that this concatenation generates TPMs in which the dominant mechanisms contributing to $\Phi$ are those consistently present across multiple responses, thereby reflecting intrinsic properties of a specific RN.

\subsubsection{Searching for Optimal Time series Satisfying Conditional Independence and Markov Property}\label{Methods_IIT_for_LLM_Representations_Searching_for_Optimal_Conditional_Independence_and_Markov_Property}

Having obtained a pool of (C)ARR, three critical questions emerged regarding the optimal concatenation of the sequences: 

\begin{enumerate}
    \item What is the ideal number of responses to select to best satisfy the conditional independence and Markov property requirements of IIT?
    \item Which specific responses should be chosen from the available pool?
    \item In what order should these selected responses be concatenated to maximize adherence to IIT principles?
\end{enumerate}

We developed a heuristic search procedure. Specifically, we searched by the total number of tokens in concatenated (C)ARR, iterating over token counts $t_i \in [50, 100, 150, 200, \cdots, 900, 950, 1000]$. In each iteration, we randomly sampled from the pool of (C)ARR, ensuring that the total number of tokens from the concatenated (C)ARR remained within $t_i$. Priority was given to the original (C)ARR, and we only sampled from the augmented ones when the target $t_i$ could not be reached. A minor adjustment was made to allow the total token count to slightly exceed $t_i$ in order to preserve the integrity of the final 
(C)ARR.

The time series of (C)ARR are characterized by a shape of $(n_{2,k}, D)$, where $D$ denotes the embedding dimension of our selected LLMs. Specifically, for LLaMA3.1-8B, Mistral-7B, and Mixtral-8x7B, $D$ is set to $4,096$. In contrast, for LLaMA3.1-70B, $D$ is $8,192$.

To address the computational complexity associated with calculating $\Phi$ values from IIT, we reduced the embedding dimension $D$ to $4$ for the concatenated (C)ARR within each iteration $i$ using Principal Component Analysis (PCA). It is important to note that the calculation of $\Phi$ scales with a complexity of $O(n53^n)$, where $n$ represents the number of nodes (which corresponds to the embedding dimension $D$ in this context). Consequently, larger networks would incur increased computational time.

The RN is a network where dimensions correspond to nodes, and the goal is to analyze the global structure and dynamics of this network over a sequence of representations. PCA excels at preserving the global structure of the data by focusing on high-variance components, which are likely to reflect the primary axes of variation in the RN. Non-linear methods, such as UMAP, may prioritize local structures at the expense of global coherence, potentially distorting the network-level properties we aim to study.

PCA remains one of the most widely used and interpretable dimensionality reduction techniques. Its linear nature allows for a straightforward projection of high-dimensional data into a reduced latent space, where each principal component is an orthogonal axis capturing maximal variance \citep{jolliffe2016principal}. This is particularly advantageous in contexts like ours, where the reduced representations are intended to serve as nodes within a graph-structured model (the RN). Unlike manifold learning techniques such as t-SNE or UMAP, PCA offers a deterministic, global projection that avoids issues of local instability and sensitivity to initialization \citep{wattenberg2016use}.

PCA has been widely used in neuroimaging and network analysis studies, including those applying IIT. For example, in the study on resting-state fMRI \citep{nemirovsky2023implementation}, dimensionality reduction was implicitly performed by clustering regions into centroids, a process that assumes linear separability akin to PCA, supporting its applicability to network-based structures like the RN. Precedent exists for using PCA to extract network-level patterns from high-dimensional neural data, including fMRI \citep{smith2015positive}, EEG \citep{blankertz2011single}, and even LLM embeddings \citep{ethayarajh-2019-contextual}. These applications demonstrate PCA's viability in compressing large-scale representations while preserving structure sufficient for meaningful downstream analysis.

While we recognize that PCA may not capture complex non-linear dynamics, our primary goal was to derive a consistent and computationally tractable embedding space suitable for the construction and analysis of RN. This choice reflects a balance between theoretical fidelity and empirical feasibility.

To determine the specific dimensionality to which the RN should be reduced using PCA, we referred to existing implementations and/or applications of IIT, such as:

\begin{itemize}
    \item \cite{albantakis2014evolution} analyzed networks with 3-6 nodes.
    \item \cite{oizumi2014phenomenology} used small networks (4–6 nodes) to demonstrate the principles of IIT, showing that even small systems can exhibit non-trivial $\Phi$ values.
    \item In the the study on implementing IIT on resting-state fMRI \citep{nemirovsky2023implementation}, resting-state networks were reduced to five regions to balance spatial resolution and computational complexity.
\end{itemize}

There are some concerns stating that information relevant to "consciousness" may reside in higher dimensions. However, PCA ensures that the reduced dimensions retain the most prominent patterns, which are likely to include the core interactions driving the RN's dynamics.

Our choice of $D = 4$ follows this precedent, ensuring that the RN is complex enough to capture meaningful interactions but simple enough for $\Phi$ computation, and serves as a practical starting point to establish the analytical framework w.r.t the RN.

IIT 3.0 \citep{oizumi2014phenomenology} and 4.0 \citep{albantakis2023integrated}, as well as the tool PyPhi (\url{https://pyphi.readthedocs.io/en/latest/}), are based on the assumption that time series adhere to the Markov property and conditional independence. Accordingly, we evaluated these assumptions for each time series obtained from the aforementioned procedures during each iteration $i$. We recorded the percent differences, denoted as $d_i$, between the Transition Probability Matrices (TPMs) derived directly from the time series and their truly conditionally independent counterparts. Additionally, we assessed the significance of deviations from the Markov hypothesis by recording the $p_i$ values, where $p_i < 0.05$ signifies a significant violation. Detailed information on these tests is available in the subsections titled "Statistical test for the Markov property" and "Test for conditional independence" under "Methods" section from \cite{nemirovsky2023implementation}. We also recorded the specific (C)ARR and their concatenation order.

It is important to note that our assessments of conditional independence and Markov property were conducted exclusively on the \textbf{binarized} (and dimensionally reduced) representation sequences, rather than on their original form. This approach mirrors the methodology employed in the analysis of \textbf{binarized} human resting-state fMRI signals reported in \cite{nemirovsky2023implementation}. This is because although IIT 3.0 and 4.0 introduce substantially more rigorous mathematical formulations than their predecessors, both $\Phi^{\max}$ (IIT 3.0) and $\Phi$ (IIT 4.0) remain restricted to discrete Markovian systems composed of binary elements \citep{oizumi2014phenomenology, albantakis2023integrated}. This constraint limits their applicability to empirical data, such as those derived from neuroimaging data or LLM representations. Nonetheless, we contend that a preliminary implementation is feasible using currently available tools. We hope this work lays a foundation for future empirical applications of IIT.

After completing all the searching iterations, we retained those cases where $p_i > 0.05$ and $d_i < 100\%$. We then identified the iteration $i$ that minimizes $d_i$, denoted as $i = \arg\min d_i$. At this stage, we obtained the optimal number of tokens from the concatenated (C)ARR, as well as the specific responses and their concatenation order.

The aforementioned search procedure was applied to \textbf{each score category of every stimulus from each ToM task, encompassing all stimuli and each linguistic span, as well as across each transformer layer from each LLM}.

However, this strategy may introduce a challenge related to the uneven distribution of total token lengths used to construct each TPM, which serves as the basis for deriving IIT estimates. First, the token lengths of an individual response vary in their own sense, even within the same score category of the same stimulus in the same ToM task. Nevertheless, token length is an inherent aspect of the content and/or degree of "experience," i.e., potentially encoded in the (C)ARR, and the IIT estimate captures both the level (degree) and qualitative aspect (qualia) of consciousness \citep{oizumi2014phenomenology, albantakis2023integrated}. In this context, variations in token length are expected, as our observations are grounded in the nature of the (C)ARR itself. Second, length differences also arise from the concatenation strategy we applied, as detailed in Sec. \ref{Methods_IIT_for_LLM_Representations_Text_Augmentation_for_Responses}. However, the effects of concatenation may be mitigated across multiple responses and are therefore intrinsic to a particular RN.

\subsubsection{Constructing the Transition Probability Matrix}\label{Methods_IIT_for_LLM_Representations_Constructing_the_Transition_Probability_Matrix}

In calculating $\Phi$, the primary input is the Transition Probability Matrix (TPM), which quantifies the probabilities of transitions between states to characterize the behavior of a system. PyPhi offers two variants of the TPM: 1) the state-by-state TPM, which details the probability that a given state of the entire network transitions to another state at a subsequent time point, with dimensions $N_{\text{states}} \times N_{\text{states}}$ (i.e., $2^4 \times 2^4$), and 2) the state-by-node TPM, which indicates the probability of a node flipping between $0$ and $1$ when the system is in a specific state, with dimensions $N_{\text{states}} \times N_{\text{nodes}}$ (i.e., $2^4 \times 4$).

Following the methodology outlined in the subsection titled "Obtaining the transition probability matrix" under the "Methods" section from \cite{nemirovsky2023implementation}, we constructed the TPM using the PyPhi tool (\url{https://pyphi.readthedocs.io/en/latest/}). Each TPM was built based on the optimal time series identified in the previous step, as described in Sec. \ref{Methods_IIT_for_LLM_Representations_Searching_for_Optimal_Conditional_Independence_and_Markov_Property}. This process was carried out for \textbf{each score category of every stimulus from each ToM task, encompassing all stimuli and linguistic spans, as well as each transformer layer from each LLM}.

\subsubsection{Estimating Integrated Information through $\mu[\Phi^{\max}]$ (IIT 3.0), $\mu[\Phi]$ (IIT 4.0), Conceptual Information $\mu[CI]$ (IIT 3.0), and $\mu[\Phi\text{-structure}]$ (IIT 4.0)}\label{Methods_IIT_for_LLM_Representations_Measuring_Integrated_Information}

\begin{enumerate}
    \item IIT 3.0: $\mu[\Phi^{\max}]$ and Conceptual Information $\mu[CI]$
\end{enumerate}

In IIT 3.0 \citep{oizumi2014phenomenology}, $\Phi^{\max}$ is the central quantity meant to capture the \textbf{amount of integrated information} in a system. It measures \textbf{how much information a system generates as a whole} beyond what is generated by its parts in isolation. This quantity is interpreted as a theoretical description of \textbf{consciousness}: a system has consciousness to the extent that it has a nonzero and irreducible $\Phi^{\max}$.

$\Phi^{\max}$ of a system is defined as the maximum of $\Phi$ over all subsystems of the system, each evaluated for their most irreducible cause-effect structure \citep{oizumi2014phenomenology}.

\begin{itemize}
    \item $\Phi$ quantifies the \textbf{irreducibility} of a mechanism's cause-effect structure.
    \item $\Phi^{\max}$ is the \textbf{maximum} $\Phi$ value across all possible \textit{subsystems} and \textit{partitions}, representing the most integrated part of the system — termed the \textbf{main complex}.
\end{itemize}

In IIT 3.0 \citep{oizumi2014phenomenology}, the Wasserstein distance, also known as the earth mover's distance (EMD), specifies the metric of concept space and thus the distance between probability distributions ($\varphi$) and between constellations of concepts ($\Phi$). The EMD used in IIT 3.0 was originally defined in \cite{rubner2000earth}.

\textbf{Conceptual Information ($CI$)} quantifies the specificity and differentiation of a mechanism's cause-effect repertoire. $CI$ is measured as the distance in concept space between $C$ — the constellation of concepts generated by a given set of elements — and $p^{uc}$, the unconstrained past and future repertoire, representing a "null" concept in which all states are equally probable in the absence of a mechanism. This distance is computed using an extended version of the EMD, which sums the standard EMD between each concept's cause-effect repertoire and $p^{uc}$, weighted by the corresponding $\varphi^{\max}$ values. The $\varphi^{\max}$ value indicates the degree to which a concept exists intrinsically.

The relationship between $\Phi^{\max}$ and $CI$:

\begin{itemize}
    \item \textbf{$CI$ measures specificity} of cause-effect relationships from a mechanism. It tells us how much information a mechanism contributes.
    \item \textbf{$\Phi$ measures irreducibility}: how much of that specificity is lost when parts are disconnected. It is the \textbf{system-level} sum of irreducible conceptual information.
    \item \textbf{$\Phi^{\max}$} is the \textbf{maximum irreducible conceptual information} generated by any subsystem of the system. It selects the \textbf{maximally integrated subsystem}, i.e., the \textit{complex}.
\end{itemize}

Following the approach of \cite{nemirovsky2023implementation}, and for the sake of computational efficiency, we evaluated only the full-network subset of each RN at specific states among the $2^4 = 16$ possible states. Since the state of the RN — and thus the value of $\Phi^{\max}$ — varies over time, we computed $\Phi^{\max}$ for every state and calculated a weighted average, denoted as $\mu[\Phi^{\max}]$. The weight assigned to each state was proportional to its frequency of occurrence within the time series.

As $\Phi^{\max}$ represents a specific instance of $\Phi$, and given that the RN is reduced to four dimensions, yielding $16$ possible mechanisms ($2^4$), the $CI$ for each mechanism is calculated accordingly:

\begin{equation}
\Phi^{\max} = \text{the particular } \Phi = \sum_{i=0}^{15} CI^{(i)} = D_{\text{e-EMD}}^{(0)} + D_{\text{e-EMD}}^{(1)} + \cdots + D_{\text{e-EMD}}^{(15)}
\end{equation}

where,

\begin{align}
CI^{(i)} = D_{\text{e-EMD}}^{(i)} \text{ is the Conceptual Information from the mechanism } i \text{ among all } 2^4 = 16 \text{ possible mechanisms.} \nonumber
\end{align}

When represented as a vector metric, the following formulation is obtained:

\begin{equation}
\Phi^{\max} \rightarrow CI = [D_{\text{e-EMD}}^{(0)}, D_{\text{e-EMD}}^{(1)}, \cdots, D_{\text{e-EMD}}^{(15)}]
\end{equation}

Analogous to the computation of $\mu[\Phi^{\max}]$ in \cite{nemirovsky2023implementation}, we calculated the weighted average of each $CI$ across the time series. Specifically, for each of the $2^4 = 16$ possible network states appearing in the time series, we counted its frequency of occurrence and used these frequencies to compute the weighted average, denoted as $\mu[CI]$.

\begin{equation}
\Phi^{\max}_j \rightarrow CI_j = [D_{\text{e-EMD}, j}^{(0)}, D_{\text{e-EMD}, j}^{(1)}, \cdots, D_{\text{e-EMD}, j}^{(15)}]
\end{equation}

\begin{equation}
\mu[CI] = \sum_{j = 0}^{15} \alpha_j \cdot CI_j
\end{equation}

where,

\begin{align}
\alpha_j \text{ is the weight of } CI_j \text{ based on the occurrence of the particular state } j \text{ among all the states in the time series.} \nonumber
\end{align}

\begin{figure}[ht!]
\centering
\includegraphics[width=1.0\columnwidth, trim={0 9.7cm 0 0},clip]{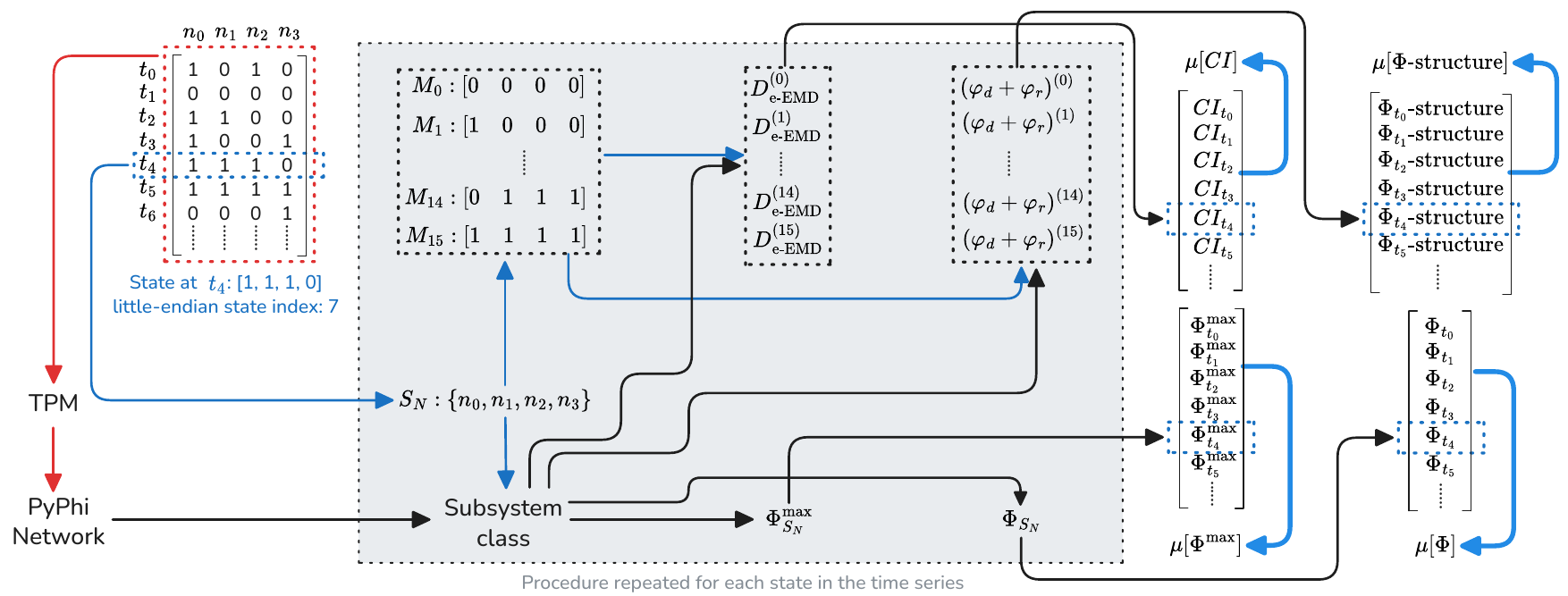}
\caption{\textbf{Summary of Procedure for Estimating $\mu[\Phi^{\max}]$ (IIT 3.0), $\mu[\Phi]$ (IIT 4.0), Conceptual Information $\mu[CI]$ (IIT 3.0), and $\mu[\Phi\text{-structure}]$ (IIT 4.0)}. Each binarized time series was used to construct a transition probability matrix (TPM) of dimensions $2^4 \times 2^4$, derived by counting the number of transitions from one state to another. For the subsystem defined by the system's state and a full subset of four nodes, we calculated $\Phi^{\max}$, $\Phi$, $CI$, and $\Phi\text{-structure}$. These metrics were computed for each state in the time series using the full subset of four nodes. The weighted averages of $\Phi^{\max}$, $\Phi$, $CI$, and $\Phi\text{-structure}$, denoted as $\mu[\Phi^{\max}]$, $\mu[\Phi]$, $\mu[CI]$, and $\mu[\Phi\text{-structure}]$, were then calculated based on the frequency of each state's occurrence in the time series.}\label{methods_computing_metrics}
\end{figure}

The procedure to obtain $\mu[\Phi^{\max}]$ and $\mu[CI]$, beginning with the time series and mechanism analysis — specifically from $M_0$ through $M_{15}$, which are used to derive the per-mechanism $CI_j$ for a given state — is summarized in Fig. \ref{methods_computing_metrics}.

Consequently, the $CI$ metric is represented as a vector with a length of $2^4 = 16$, corresponding to the number of RN states. $CI$ serves as a decomposition of the scalar metric $\Phi^{\max}$.

\begin{enumerate}
    \setcounter{enumi}{1}
    \item IIT 4.0: $\mu[\Phi]$ and $\mu[\Phi\text{-structure}]$
\end{enumerate}

$\Phi$ in IIT 4.0 \citep{albantakis2023integrated} quantifies the \textbf{integrated cause–effect power} of a physical system in a particular state. It is not just about information content, but about information \textbf{intrinsically} specified by the system about itself — i.e., how the system's current state constrains its \textbf{own past and future states} in a way that is irreducible to parts.

A \textbf{$\Phi\text{-structure}$} is the full \textbf{cause–effect structure} specified by a system when $\Phi > 0$. It is made up of:

\begin{itemize}
    \item \textbf{Distinctions}: Mechanisms that specify irreducible cause–effect repertoires.
    \begin{itemize}
        \item A \textbf{distinction} is specified by a mechanism in a particular state that has \textbf{irreducible intrinsic cause–effect power}.
        \item Formally: If a mechanism $M$ in state $m$ has $\varphi > 0$, it forms a \textbf{distinction}.
        \item The distinction includes:
        \begin{itemize}
            \item The mechanism $M$;
            \item Its state $m$;
            \item Its integrated cause and effect repertoires.
        \end{itemize}

    \end{itemize}
    \item \textbf{Relations}: Higher-order relations among distinctions (how their repertoires relate and overlap).
    \begin{itemize}
        \item Relations are \textbf{higher-order structures} that capture \textbf{how distinctions constrain each other}.
        \item Defined based on \textbf{overlap and mutual compatibility} between the cause–effect repertoires of distinctions.
        \item Relations are not just semantic or functional connections but are intrinsic and structured by causal power.
    \end{itemize}
\end{itemize}

Distinctions and relations together form the $\Phi\text{-structure}$. According to IIT 4.0 \citep{albantakis2023integrated}, the \textbf{qualia} — the contents and structure of conscious experience — are identified with the \textbf{shape} of this \textbf{$\Phi\text{-structure}$}.

Following \cite{nemirovsky2023implementation}, and for the sake of computational efficiency, we evaluated only the full-network subset of each RN at specific states among the $2^4 = 16$ possible states. Since the state of the RN — and consequently the value of $\Phi$ — varies throughout the time series, we calculated $\Phi$ for each state and derived a weighted average, denoted as $\mu[\Phi]$. The weight assigned to each state was proportional to its frequency of occurrence within the time series.

Given the reduction of the RN's dimensionality to four, resulting in $2^4 = 16$ possible mechanisms, the calculation of each $\Phi\text{-structure}$ proceeds as follows:

\begin{equation}
\Phi = \sum_{\text{distinctions}} \varphi_d + \sum_{\text{relations}} \varphi_r = \sum_{i=0}^{15} \Phi^{(i)}\text{-structure} = (\varphi_d + \varphi_r)^{(0)} + (\varphi_d + \varphi_r)^{(1)} + \cdots + (\varphi_d + \varphi_r)^{(15)}
\end{equation}

where,

\begin{align}
\Phi^{(i)}\text{-structure} = (\varphi_d + \varphi_r)^{(i)} \text{ is the $\Phi\text{-structure}$ from the mechanism } i \text{ among all } 2^4 = 16 \text{ possible mechanisms.} \nonumber
\end{align}

When expressed as a vector metric, the formulation is given by:

\begin{equation}
\Phi \rightarrow \Phi\text{-structure} = [(\varphi_d + \varphi_r)^{(0)}, (\varphi_d + \varphi_r)^{(1)}, \cdots, (\varphi_d + \varphi_r)^{(15)}]
\end{equation}

Following \cite{nemirovsky2023implementation}, for each state — among the $2^4 = 16$ possible states — appearing in the time series, we counted its occurrences and calculated the weighted average of each $\Phi\text{-structure}$ across the time series, denoted as $\mu[\Phi\text{-structure}]$.

\begin{equation}
\Phi_j \rightarrow \Phi_j\text{-structure} = [(\varphi_d + \varphi_r)_j^{(0)}, (\varphi_d + \varphi_r)_j^{(1)}, \cdots, (\varphi_d + \varphi_r)_j^{(15)}]
\end{equation}

\begin{equation}
\mu[\Phi\text{-structure}] = \sum_{j = 0}^{15} \alpha_j \cdot \Phi_j\text{-structure}
\end{equation}

where,

\begin{align}
\alpha_j \text{ is the weight of } \Phi_j\text{-structure} \text{ based on the occurrence of the particular state } j \text{ among all the states in the time series.} \nonumber
\end{align}

The procedure for calculating $\mu[\Phi]$ and $\mu[\Phi\text{-structure}]$, beginning with the time series and mechanism analysis ($M_0$ to $M_{15}$) to derive the per-mechanism $\mu[\Phi_j\text{-structure}]$ for each state, is summarized in Fig. \ref{methods_computing_metrics}.

Consequently, the $\Phi$-structure metric is also represented as a vector of length $2^4 = 16$, corresponding to the total number of distinct RN states. The $\Phi$-structure, like the $CI$, serves as a decomposition of the scalar metric $\Phi$.

We computed the weighted average for all the four metrics, resulting in $\mu[\Phi^{\max}]$ (IIT 3.0), $\mu[\Phi]$ (IIT 4.0), Conceptual Information $\mu[CI]$ (IIT 3.0), and $\mu[\Phi\text{-structure}]$ (IIT 4.0), for \textbf{each score category of every stimulus from each ToM task}. This analysis covered \textbf{all stimuli and linguistic spans, as well as each transformer layer from every LLM}. 

\subsubsection{Systematic Spatio-temporal Permutation}\label{Methods_IIT_for_LLM_Representations_Systematic_Spatio_temporal_Permutation}

The core of our data analysis involves LLM representations, processed as the RN and treated as time series. The nodes for such RN correspond to the learned features spanning across the embedding dimension of each representation, and we assume the RN is fully connected, with potential latent connections (edges) between nodes or node clusters. Inspired by \citep{nemirovsky2023implementation} we developed control procedures for spatio-temporal permutations for these representations.

Arbitrarily permuting embeddings derived from LLMs is generally discouraged, as these representations encode critical features learned during training. The dimensional order of these vector representations is essential, maintaining semantic integrity and compatibility with downstream computational processes. However, our study requires permutation along the embedding dimension for each (C)ARR, based on the conceptualization of learned features in the vector as nodes within our proposed RN. Drawing inspiration from \cite{nemirovsky2023implementation}, we hypothesize latent interconnections (edges) between individual nodes or node clusters. Consequently, spatial permutations facilitate a systematic investigation of these potential nodal relationships within the RN.

For the spatio control procedure, we randomly permuted the time series along the embedding dimension for each LLM to randomize the arrangements of the embedding order. This procedure was repeated $10$ times for \textbf{each score category of every stimulus from each ToM task, covering all stimuli and linguistic spans, as well as each transformer layer from each LLM}.

For the temporal control procedure, we preserved the original order of each token embedding but permuted the sequence in which the (C)ARR were concatenated. This process was repeated $10$ times for \textbf{each score category of every stimulus from each ToM task, encompassing all stimuli and linguistic spans, as well as each transformer layer from each LLM}.

Both control procedures were integrated into the search for the optimal time series that satisfy conditional independence and the Markov property, as detailed in Sec. \ref{Methods_IIT_for_LLM_Representations_Searching_for_Optimal_Conditional_Independence_and_Markov_Property}.

\subsection{Deriving Span Representations from LLM Representations}\label{Methods_Span_Representations_from_LLM_Representations}

The integrated information derived from LLM representations is one of the factors we use to interpret ToM test results. But it is insufficient to give a complete explanation to the differences from ToM test performances as discussed in Sec. \ref{Introduction}. We further explored whether these results can be attributed to other factors other than IIT estimates, such as the Span Representations, using the same (dimensionality reduced) LLM representations. This investigation aims to distinguish between phenomena that might suggest consciousness and (well) separated manifolds inherent to LLM representations.

The study by \cite{peters-etal-2018-dissecting} suggests that the context vectors of bi-directional language models undergo abrupt changes at syntactic boundaries, allowing them to be utilized for constructing representations of spans or consecutive token sequences.

In fact, this method also effectively captures span-level information for non-bidirectional language models, i.e.,

\begin{itemize}
    \item \textbf{Capturing Boundary Information}: The method relies on the context vectors at the span's boundaries ($\mathbf{h}_{s_0, i}$, $\mathbf{h}_{s_1, i}$). In non-bidirectional models, $\mathbf{h}_{s_0, i}$ encodes information from all tokens up to $s_0$, and $\mathbf{h}_{s_1, i}$ encodes information up to $s_1$. While these vectors lack future context (unlike biLMs), they still summarize the sequence's state at these points. This makes them suitable anchors for defining a span, as they reflect the cumulative context leading into and through the span. Studies like \cite{tenney2019you} show that unidirectional models encode significant positional and sequential information, supporting the use of boundary vectors for span representation.
    \item \textbf{Element-Wise Product ($\mathbf{h}_{s_0, i} \odot \mathbf{h}_{s_1, i}$)}: The product captures interactions between the boundary vectors, highlighting shared or amplified features. In non-bidirectional models, this operation can emphasize common contextual patterns (i.e., syntactic or semantic consistencies) between the start and end of the span, even if the context is unidirectional. For example, if a span covers a noun phrase, the product may highlight shared lexical or structural features encoded up to each boundary.
    \item \textbf{Difference ($\mathbf{h}_{s_0, i} - \mathbf{h}_{s_1, i}$)}: The difference captures changes in the contextual representation from the start to the end of the span. In unidirectional models, $\mathbf{h}_{s_1, i}$ includes the influence of tokens between $s_0$ and $s_1$, so the difference reflects the span's contribution to the evolving context. This is particularly useful for identifying shifts in meaning, topic, or structure within the span, even without bidirectional sensitivity. \cite{ethayarajh-2019-contextual} notes that unidirectional models produce embeddings that vary smoothly across tokens, implying that the difference operation can still capture meaningful transitions within a span.
    \item \textbf{Generalizability to Unidirectional Context}: The method does not explicitly require bidirectional context; it operates on any pair of contextualized vectors. Non-bidirectional models like GPT generate rich, context-sensitive embeddings through their transformer layers, as shown in \cite{brown2020language}. These embeddings encode sufficient information about sequence structure, allowing the concatenation of boundary vectors, their product, and difference to form a robust span representation. The method's design leverages relative positional information (start vs. end of span), which aligns with how unidirectional models process sequences incrementally.
\end{itemize}

Following the methodology outlined in \cite{jawahar-etal-2019-bert}, we converted the LLM representations of interest for all sampled (C)ARR, across both spatio-temporal permutation control groups described in Sec. \ref{Methods_IIT_for_LLM_Representations_Systematic_Spatio_temporal_Permutation}, into span representations according to \cite{peters-etal-2018-dissecting}.

Specifically, for a span of $S$ tokens from indices $s_0$ to $s_1$, we calculated a Span Representation, $\mathbf{s}_{(s_0,s_1),i}$, at layer $i$. This representation was constructed by concatenating the first token representation ($\mathbf{h}_{s_0, i}$), the last token representation ($\mathbf{h}_{s_1, i}$), their element-wise product, and difference:

\begin{equation}
\mathbf{s}_{(s_0,s_1),i} = [\mathbf{h}_{s_0, i}; \mathbf{h}_{s_1, i}; \mathbf{h}_{s_0, i} \odot \mathbf{h}_{s_1, i}; \mathbf{h}_{s_0, i} - \mathbf{h}_{s_1, i}] \label{span_representations}
\end{equation}

\subsection{Statistical Analysis}\label{Methods_Statistical_Analysis}

\subsubsection{Criterion 1 based on $\Phi$ Value Comparisons}\label{Methods_Phi_Value_Comparisons}

\textbf{Granularity at the Stimulus Level, w/ Aggregation}: Following the methodology outlined in Sec. \ref{Methods_Phi_Value_Distributions}, we further aggregated the samples under the $10$ randomization controls (Sec. \ref{Methods_IIT_for_LLM_Representations_Systematic_Spatio_temporal_Permutation}). Each sample was grouped by the same LLM, ToM Task, Sheet, and Stimulus. For comparisons across different transformer layers, we also ensured that the same Transformer Layer was used. Similarly, for comparisons across different linguistic spans, the same Transformer Layer along with the same linguistic span was considered. The results were visualized as line graphs with shaded ribbon areas, where the solid line represents the mean values, and the ribbon shows the range from the lowest to the highest $\Phi$ values across the $10$ randomization results. An additional filter was applied such that each score category for each sample must include at least one valid $\Phi$ estimate from the $10$ results. To assess statistical differences between score categories, a one-way Wilcoxon test was performed on the filtered samples, and the corresponding $p$ values were annotated on the line graphs. For the ToM Task with three score categories, Holm corrections were applied to the $p$ values in pairwise comparisons (i.e., Score $0$ vs. Score $1$, Score $0$ vs. Score $2$, and Score $1$ vs. Score $2$).

We transformed the line graph into a stacked bar representation, categorizing results as "good" or "bad" based on the averaged $\Phi$ estimates across different score categories. For two-score tasks, a result was classified as "good" when the $\Phi$ estimate for Score $1$ exceeded that of Score $0$, with the inverse defining "bad" cases. In three-score ToM Task, "good" was defined as Score $2$ $\ge$ Score $1$ $\ge$ Score $0$, with deviations from this pattern classified as "bad". (\textbf{Criterion 1}) We established a significance threshold where "good" cases must comprise over $80\%$ of valid stimuli for each ToM Task (see also the decomposition of our primary research question in Sec. \ref{Introduction}).

\subsubsection{Criterion 2 based on $\Phi$ Value Distributions}\label{Methods_Phi_Value_Distributions}

\textbf{Granularity at the ToM Task Level, w/o Aggregation}: We grouped the values of $\mu[\Phi^{\max}]$ (IIT 3.0) and $\mu[\Phi]$ (IIT 4.0) across different ToM Tasks, LLMs, transformer layers, and linguistic spans, and visualized them using violin and box plots. To assess the statistical significance, we conducted a two-way Wilcoxon test between samples from different score categories and annotated the resulting $p$ values on top of each pair of Score 0 and Score 1 in the violin plots. The Holm corrections were also applied to the $p$ values for the ToM Task with three score categories: Score 0 vs. Score 1, Score 0 vs. Score 2, and Score 1 vs. Score 2.

\textbf{Criterion 2}: A degree of statistical significance ($p < 0.05$) is expected if $\mu[\Phi^{\max}]$ (IIT 3.0) and/or $\mu[\Phi]$ (IIT 4.0) can effectively distinguish between different ToM Task score categories (see also the decomposition of our primary research question in Sec. \ref{Introduction}).

\subsubsection{Criterion 3 based on Comparing Interpretation Abilities across Different Metrics}\label{Methods_Comparing_Interpretation_Abilities_across_Different_Metrics}

\textbf{Granularity at the ToM Task Level, w/o Aggregation}: To compare various metrics, including $\mu[\Phi^{\max}]$ (IIT 3.0), $\mu[CI]$ (IIT 3.0), $\mu[\Phi]$ (IIT 4.0), $\mu[\Phi\text{-structure}]$ (IIT 4.0), and Span Representations, we applied logistic regression and calculated the area under the receiver operating characteristic (ROC) curve (AUC). For ToM tasks with three categorical scores, such as the Strange Stories task, we used multinomial logistic regression in combination with the one-vs-rest multiclass ROC AUC. To train each valid logistic regression model, we required that the number of samples for both Score $0$ and Score $1$ (or for Scores $0$, $1$, and $2$ in ToM task with three scores) be at least equal to the number of features (i.e., $\geq 16$). This filtering criterion was applied in addition to the procedure outlined in Sec. \ref{Methods_Phi_Value_Distributions}. The mean AUC values from 5 iterations of 5-fold cross-validation were computed and presented as bar graphs with error bars representing standard deviation ($\pm$). A maximum of $10,000$ iterations was set for model convergence.

\textbf{Criterion 3}: We define the mean AUC for each metric as an indicator of the interpretative capability of the ToM test results. A metric with a higher mean AUC suggests that the ToM test result is more strongly attributed to that metric (see also the decomposition of our primary research question in Sec. \ref{Introduction}).

\section{Results}\label{Results}

\subsection{Results Overview}\label{Results_Overview}

A total of $165,365$ valid samples were obtained through our data processing pipeline, as described in Sec. \ref{Methods_IIT_for_LLM_Representations} and Sec. \ref{Methods_Span_Representations_from_LLM_Representations}. The properties associated with each sample are summarized in Tables \ref{Table_Results_Overview} and \ref{Table_Results_Overview_Contd}.

\begin{sidewaystable}[]
\caption{Results Overview}\label{Table_Results_Overview}
\begin{tabular}{llll}
\toprule
Property  & Value & Comment & Reference\\
\midrule

Linguistic Span & \textit{Entire}, \textit{Complement}, or \textit{MSV} & The specific linguistic range for the stimulus upon & Sec. \ref{Methods_IIT_for_LLM_Representations_Response_Representations_Attended_to_Stimulus_Linguistic_Spans_in_Context} and Sec. \ref{Methods_IIT_for_LLM_Representations_Response_Representations_Attended_to_Stimulus_Linguistic_Spans_in_Context} \\
 &  & which the consciousness estimate, derived from  & \\
 &  & the IIT, is based. & \\

Model Name & \textit{LLaMA3.1-8B}, \textit{LLaMA3.1-70B} &  & Appendix A.1 \\
 & \textit{Mistral-7B}, or \textit{Mixtral-8x7B} &  &  \\

Transformer Layer & \textit{$0$ - $11$}, or \textit{$\frac{2}{3}$} & The specific transformer layer for the stimulus upon & Sec. \ref{Methods_IIT_for_LLM_Representations_Representations_from_Multiple_Transformer_Layers} \\
 &  & which the consciousness estimate, derived from & \\
 &  & the IIT, is based. & \\

ToM Task & \textit{Hinting}, \textit{False Belief}, \textit{Irony} &  & Appendix A.2 \\
 & \textit{Strange Stories (2 scores)}, or &  &  \\
 & \textit{Strange Stories (3 scores)} &  &  \\

Sheet & i.e., \textit{ToM-B Hinting} & The particular sheet name from the original  & \\
 &  & dataset (\url{https://osf.io/dbn92}). & \\

Stimulus & i.e., \textit{Answer1} & The particular question from the ToM task.  & \\

Score & $0$, $1$, or $2$ & The score rating (category) for the response.  & \\

Limited \# Tokens & i.e., $550$ & The limited number of tokens $t_i \in [50, 100, \cdots, 950, 1000]$   & Sec. \ref{Methods_IIT_for_LLM_Representations_Searching_for_Optimal_Conditional_Independence_and_Markov_Property} \\
 &  & during each iteration from the search procedure. & \\

Actual \# Tokens & i.e., $557$ & The actual number of tokens, which might slightly exceed & Sec. \ref{Methods_IIT_for_LLM_Representations_Searching_for_Optimal_Conditional_Independence_and_Markov_Property} \\
 &  & $t_i$ during each iteration from the search procedure. & \\

Permutation & \textit{Temporal} or \textit{Spatio} & The system level randomization control.  & Sec. \ref{Methods_IIT_for_LLM_Representations_Systematic_Spatio_temporal_Permutation} \\
Control &  & & \\

Seed & i.e., $42$ & The randomization seed used in the control & Sec. \ref{Methods_IIT_for_LLM_Representations_Systematic_Spatio_temporal_Permutation} \\
 &  & procedure ranges from $42$ to $51$ for \textit{temporal} & \\
 &  & permutation, and again from $42$ to $51$ for  & \\
 &  & \textit{spatio} permutation. & \\

\bottomrule
\end{tabular}

\end{sidewaystable}

\begin{sidewaystable}[]
\caption{Results Overview (Cont'd)}\label{Table_Results_Overview_Contd}
\begin{tabular}{llll}
\toprule
Property  & Value & Comment & Reference\\
\midrule

$\Phi^{\max}$ Value (IIT 3.0) & A scalar value, i.e., $0.37$  & The estimated $\mu[\Phi^{\max}]$ (IIT 3.0).  & Sec. \ref{Methods_IIT_for_LLM_Representations_Measuring_Integrated_Information} \\

$\Phi$ Value (IIT 4.0) & A scalar value, i.e., $8.3$  & The estimated $\mu[\Phi]$ (IIT 4.0).  & Sec. \ref{Methods_IIT_for_LLM_Representations_Measuring_Integrated_Information} \\

Conceptual Information (CI) (IIT 3.0) & A vector with length $2^4=16$  & The estimated $\mu[CI]$ (IIT 3.0).  & Sec. \ref{Methods_IIT_for_LLM_Representations_Measuring_Integrated_Information} \\

$\Phi$-structure (IIT 4.0) & A vector with length $2^4=16$  & The estimated $\mu[\Phi\text{-structure}]$ (IIT 4.0).  & Sec. \ref{Methods_IIT_for_LLM_Representations_Measuring_Integrated_Information} \\

Span Representation &  & The acquired metric, aside from the estimates for consciousness.  & Sec. \ref{Methods_Span_Representations_from_LLM_Representations} \\

\bottomrule
\end{tabular}

\end{sidewaystable}

IIT 3.0 \citep{oizumi2014phenomenology} and IIT 4.0 \citep{albantakis2023integrated}, along with the PyPhi toolkit (\url{https://pyphi.readthedocs.io/en/latest/}), rely on the assumptions that time series satisfy the Markov property and conditional independence. All time series for valid samples adhering to these properties after being filtered through the search procedure (Sec. \ref{Methods_IIT_for_LLM_Representations_Searching_for_Optimal_Conditional_Independence_and_Markov_Property}). However, in several cases, particularly those involving the LLM \textit{Mixtral-8x7B}, we were unable to initialize a valid network from the TPM using PyPhi under the IIT 4.0 framework. These samples were consequently excluded and marked as invalid.

\subsection{$\Phi$ Value Distributions and Comparisons across ToM Tasks and LLMs}\label{Phi_Distribution_and_Comparison_across_ToM_Tasks_andLLMs}

At the core of IIT, $\Phi$ quantifies the consciousness of a system. A network with a higher $\Phi$ value is presumed to be with more integrated information than one with a lower $\Phi$, provided both networks are analyzed under the same contextual conditions \citep{oizumi2014phenomenology, albantakis2023integrated}. Accordingly, a higher score on a ToM test response from a specific conscious subject — an adult (aged 18–70), native English speaker, with no history of psychiatric conditions, particularly dyslexia \citep{strachan2024testing} — is expected to yield higher $\Phi$ estimates if "consciousness" phenomenon can be observed from the ToM-related "brain" network, i.e., the RN derived from (C)ARR, as discussed in Sec. \ref{Materials and Methods Overview}.

\begin{figure}[ht!]
\centering
\includegraphics[width=1.0\columnwidth, trim={0 0cm 0 0},clip]{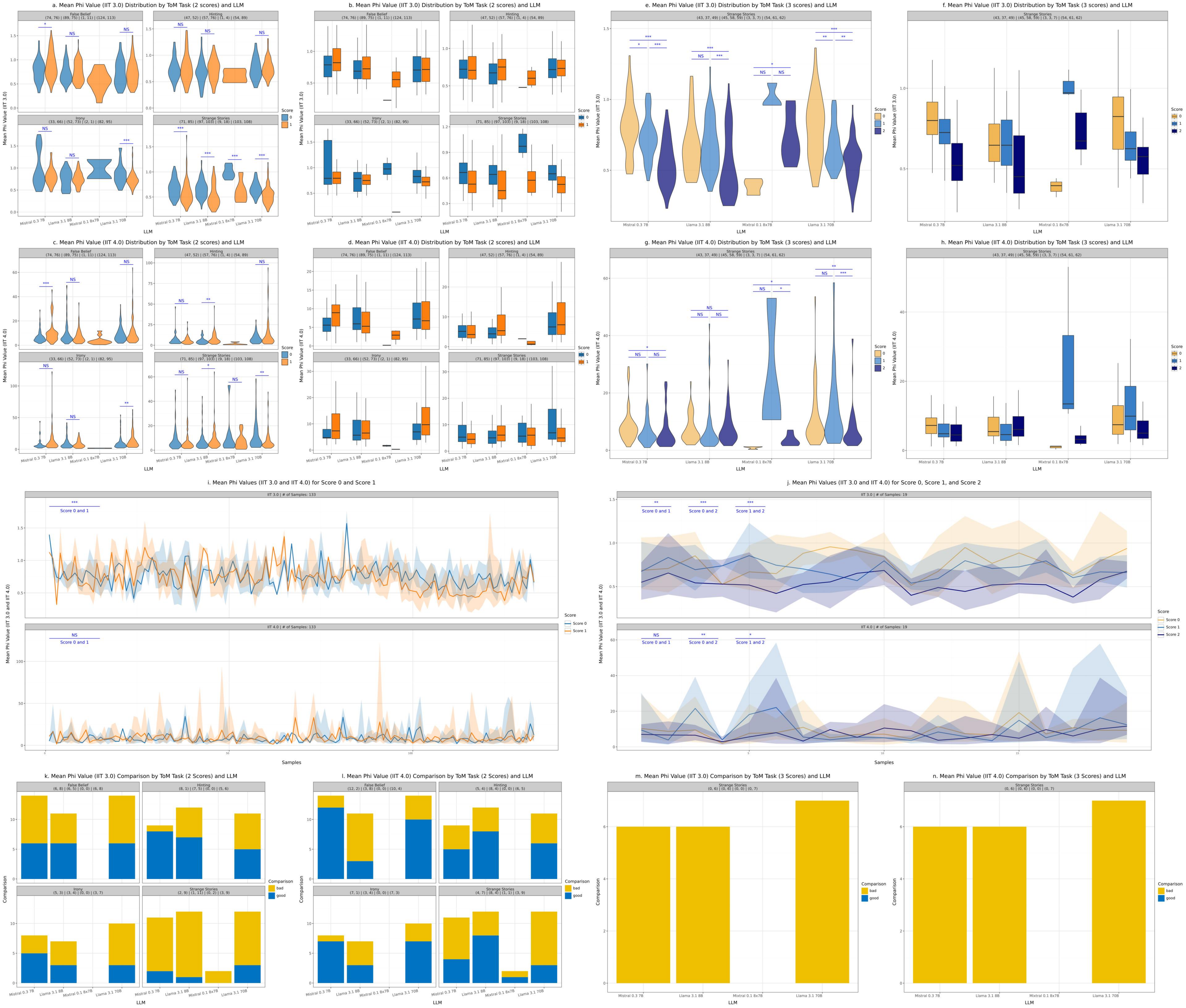}
\caption{\textbf{$\Phi$ Value Distributions and Comparisons.} \textbf{(a)} Violin plots for $\mu[\Phi^{\max}]$ (IIT 3.0) distributions by 4 ToM Tasks (2 scores) across 4 LLMs. The annotation on top of each pair of 2 scores indicates the significance ($p$ value) for Wilcoxon two-way test between the two. \textbf{(b)} The distributions presented in box plots based on the same data from \textbf{(a)}. \textbf{(c)} Similar violin plots to \textbf{(a)} but for $\mu[\Phi]$ (IIT 4.0). \textbf{(d)} The distributions presented in box plots based on the same data from \textbf{(c)}. \textbf{(e)} Violin plot for $\mu[\Phi^{\max}]$ (IIT 3.0) distributions by 1 ToM Task (3 scores) across 4 LLMs. The annotation on top of each pair of 2 scores indicates the the significance ($p$ value) for Holm-corrected Wilcoxon two-way tests between the two. \textbf{(f)} The distributions presented in box plots based on the same data from \textbf{(e)}. \textbf{(g)} Similar violin plots to \textbf{(e)} but for $\mu[\Phi]$ (IIT 4.0). \textbf{(h)} The distributions presented in box plots based on the same data from \textbf{(g)}. \textbf{(i)} Line graphs with ribbon regions for the grouped $\Phi$ value distributions (IIT 3.0 and 4.0) and filtered with ToM Tasks with 2 scores, the annotation indicates the significance ($p$ value) for Wilcoxon one-way test between the two scores on each sub-plot. \textbf{(j)} Similar Line graphs with ribbon regions to \textbf{(i)} except that the ToM Task is with 3 scores, Holm corrections are also added to the significance ($p$ value) for Wilcoxon one-way test between each pair of two scores. \textbf{(k)} Stacked bar plots based on the same data in \textbf{(i)} (IIT 3.0 only). \textbf{(l)} Similar stacked bar plots to \textbf{(k)}, but for IIT 4.0 only. \textbf{(m)} Stacked bar plots based on the same data in \textbf{(j)} (IIT 3.0 only). \textbf{(n)} Similar stacked bar plots to \textbf{(m)}, but for IIT 4.0 only. The exact number of samples are provided under the title for each sub-plot (Those for \textbf{(i)} and \textbf{(j)} are shown for aggregated samples only).}\label{main_result_fig_1}
\end{figure}

\textbf{Granularity at ToM Task Level, w/o Aggregation}: Sub-figs (a)–(d) in Fig. \ref{main_result_fig_1} depict the $\Phi$ value distributions for all four ToM tasks (2 scores) across the four LLMs, derived from the valid samples (Sec. \ref{Results_Overview}) without additional filtering. Following standard practice, the last transformer layer was used to represent the LLM. By default, the linguistic span for each sample was set to encompass the entire stimulus. Additionally, the embedding order for each ARR was preserved without \textit{spatio} permutations. Sub-figs (e) and (f) in Fig. \ref{main_result_fig_1} present analogous distributions but for the Strange Stories, which involves 3 scores.

\textbf{Granularity at the Stimulus Level, w/ Aggregation}: Sub-fig (i) in Fig. \ref{main_result_fig_1} depicts a filtered and aggregated scenario where each sample is grouped under the $10$ randomization controls (\ref{Methods_IIT_for_LLM_Representations_Systematic_Spatio_temporal_Permutation}). The solid line indicates the averages, while the shaded ribbon represents the range of $\Phi$ values, spanning from the lowest to the highest across the $10$ results. Each sample is constrained to the same LLM, ToM Task, Sheet, and Stimulus (Sec. \ref{Results_Overview}) to ensure a fair and precise comparison. Additionally, we require at least one valid $\Phi$ estimate from the $10$ results for both Score 0 and Score 1 for each sample. Sub-fig (j) in Fig. \ref{main_result_fig_1} illustrates the corresponding analysis for the Strange Stories, which involves 3 scores.

Correspondingly, the results shown in sub-fig (i) of Fig. \ref{main_result_fig_1} are further decomposed into sub-figs (k) and (l), categorized by ToM Tasks across LLMs. A result is counted as "good" if the average $\Phi$ estimate for Score $1$ exceeds that for Score $0$. Conversely, it is counted as "bad" if otherwise. Similarly, for ToM Task with three scores, the criterion is defined as "good" if the average $\Phi$ estimates satisfy Score $2$ $\ge$ Score $1$ $\ge$ Score $0$, and "bad" otherwise, as shown in sub-figs (m) and (n) of Fig. \ref{main_result_fig_1}.

\textbf{Criterion 1}: In rare instances, such as Mistral-7B on Hinting (IIT 3.0), Mistral-7B on False Belief (IIT 4.0), and Mistral-7B on Irony (IIT 4.0), the number of "good" cases exceeds the number of "bad" ones to a certain extent, as shown in sub-figs (k) and (l). However, it is important to note that the number of valid samples represented in the bar plots is limited: the total number of valid stimuli (the details are provided in Appendix A.2) is 13 for Hinting, 19 for False Belief, and 12 for Irony. Notably, none of these cases satisfies the criterion of "good" cases constituting more than $80\%$ of the total valid stimuli for each ToM Task, which undermines the statistical significance of such comparisons. Furthermore, no such cases are identified for ToM Tasks involving three scores.

\textbf{Criterion 2}: We anticipate statistical significance in the differentiation of $\Phi$ values across score categories, as assessed by Wilcoxon two-way tests (the Holm-correction is applied to adjust the $p$ values for analyses involving three score categories). Among the previously discussed cases, Mistral-7B on False Belief (IIT 4.0) emerges as the sole instance exhibiting statistically significant differences, as demonstrated in sub-fig (c) of Fig. \ref{main_result_fig_1}.

\subsection{Comparing Metrics Estimating Consciousness and Other Independent Metric}\label{Comparing_Metrics_Estimating_Consciousness_and_Other_Independent_Metric}

The distribution of "Metric-Score" can also be formulated as a classification task, where metrics serve as features and scores as targets. Within this framework, we further investigated which metric most effectively explains its corresponding score. Additionally, a new metric, Span Representation, was included in the comparison. Span Representation characterizes the span-level information of sequence of representations \citep{peters-etal-2018-dissecting, jawahar-etal-2019-bert} and is independent of any estimate of consciousness. The dimensionality-reduced (C)ARR used to derive the Span Representation is identical to that used to generate the corresponding RN and, subsequently, all IIT estimates. This restriction ensured both fairness and precision in the comparison.

\begin{figure}[ht!]
\centering
\includegraphics[width=1.0\columnwidth, trim={0 0cm 0 0},clip]{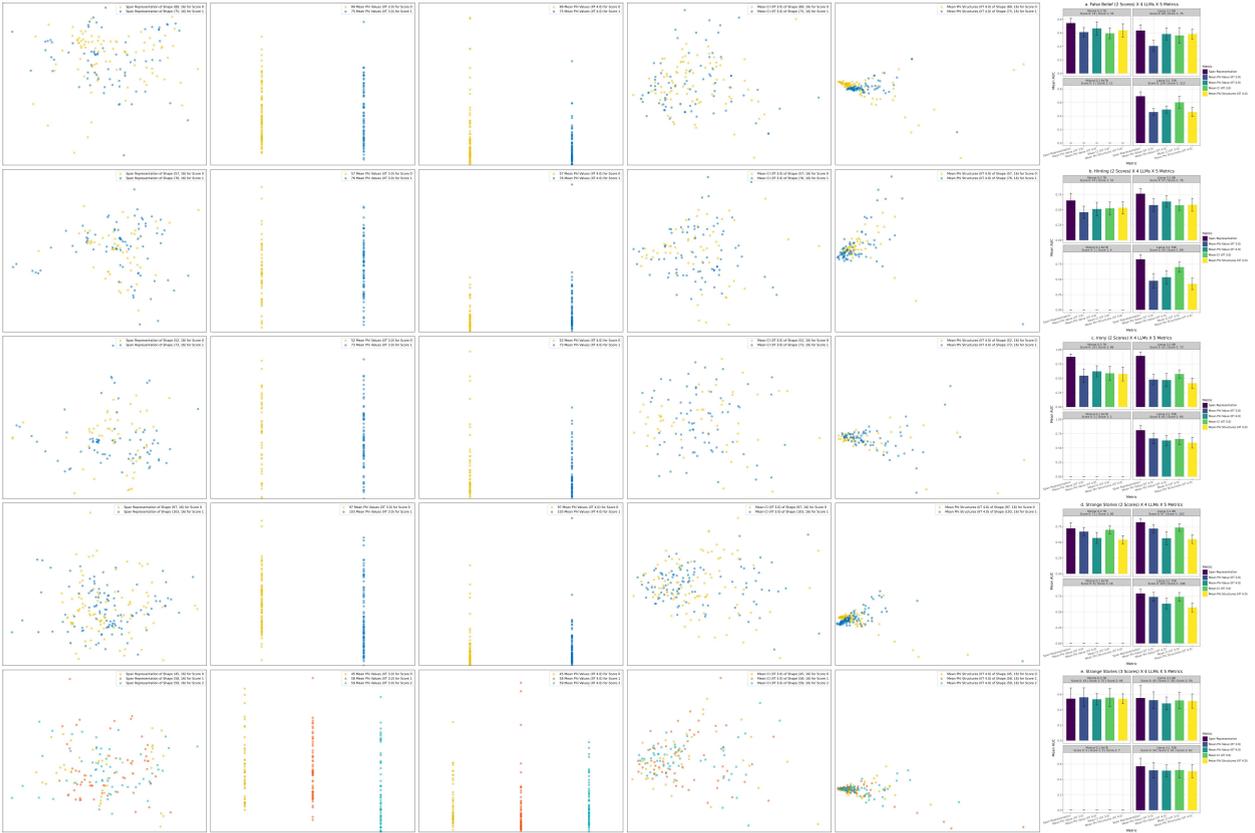}
\caption{\textbf{Scatter Plots and Mean AUCs for Different Metrics.} The scatter plots in the first five columns show the distributions of all five metrics (projected onto $2$-D space) across the four ToM Tasks. The last (fifth) row also presents data for the ToM Task Strange Stories, but with three scores. Each ToM Task is represented by a separate row. For demonstration purposes, we present only the results from the LLM \textit{LLaMA3.1-8B} in the scatter plots. The mean AUCs, derived from predictions on each test set through 5-time 5-fold cross-validated logistic regression, are shown in the final (sixth) column: \textbf{(a)} False Belief, \textbf{(b)} Hinting, \textbf{(c)} Irony, \textbf{(d)} Strange Stories ($2$ scores), and \textbf{(e)} Strange Stories ($3$ scores). The error bars, representing $\pm$ standard deviation, are included for each mean AUC. The exact number of samples for each sub-plot is provided under its title.}\label{main_result_fig_2}
\end{figure}

The scatter plots in the first five columns of Fig. \ref{main_result_fig_2} depict the distributions of all five metrics, when projected onto a $2$-D space, across the four ToM tasks. The fifth row corresponds to the ToM Task Strange Stories, but with three scores. Each row in the figure represents a different ToM task. For brevity, only the results from the LLM \textit{LLaMA3.1-8B} are shown in the scatter plots.

\textbf{Granularity at ToM Task Level w/o Aggregation}: Detailed comparisons between different metrics, specifically the mean AUCs derived from predictions on each test set during 5-time 5-fold cross-validated logistic regression analyses, are presented in the final (6th) column of Fig. \ref{main_result_fig_2}. To ensure the validity of each logistic regression model, we required that the number of samples for both Score $0$ and Score $1$ (or Score $0$, $1$, and $2$ for ToM Task with three scores) was at least equal to the number of features, i.e., 16, per sample. This additional filtering criterion was applied subsequently.

\textbf{Criterion $3$}: The detailed comparisons consistently demonstrate that the Span Representation outperforms all consciousness estimates derived from IIT in terms of mean AUC. The sole exception is a single case involving Mistral-7B on Strange Stories (three scores), where still none of the IIT-derived estimates for consciousness achieve a statistically significant improvement over the Span Representation.

In other words, the variations in ToM Task test score categories are more likely attributed to the span-level information of the LLM representation sequence rather than to a "consciousness" phenomenon as suggested by IIT estimates.

\subsection{Results Extended to Different Transformer Layers}\label{Results_Extended_to_Different_Transformer_Layers}

\footnote{All the figures in this section are provided in Appendix A.7.1.}Our study reveals quantitative distinctions in consciousness estimates across different Transformer Layers. By applying the same settings as in Sec. \ref{Phi_Distribution_and_Comparison_across_ToM_Tasks_andLLMs}, with an additional filter based on the transformer layer, Appendix Fig. 5 through Appendix Fig. 8 illustrate the distributions of $\Phi$ values across all 13 transformer layers (the 12 evenly sampled layers from each LLM, plus the $\frac{2}{3}$ layer; see Sec. \ref{Methods_IIT_for_LLM_Representations_Representations_from_Multiple_Transformer_Layers} for details). Appendix Fig. 9 presents similar results, except for the ToM Task Strange Stories with three scores. The distribution of $\Phi$ values, grouped by the 10 randomization controls, is shown in Appendix Fig. 10, with Appendix Fig. 11 corresponding to the ToM Task Strange Stories with three scores. We further decompose the results from Appendix Fig. 10 in Appendix Fig. 12 and Appendix Fig. 13. The "good" and "bad" statistics carry the same meaning as defined in Sec. \ref{Phi_Distribution_and_Comparison_across_ToM_Tasks_andLLMs}. The results for the ToM Task with three scores are shown in Appendix Fig. 14. Finally, the interpretation ability of various metrics, including the Span Representation, which is independent of any estimate of consciousness, is presented in Appendix Fig. 15 (with Appendix Fig. 16 corresponding to the ToM Task with three scores).

\textbf{Criterion 1}: The following cases, although rare, exhibit a significantly greater number of "good" cases than "bad" ones (see Sec. \ref{Methods_IIT_for_LLM_Representations_Systematic_Spatio_temporal_Permutation} for the definition of this significance).

\begin{enumerate}
    \item Layer $24$ (indexed at $8$) of LLaMA3.1-8B on Hinting (IIT 3.0 and 4.0) (Sub-figs (c) and (d) in Appendix Fig. 12)
    \item Layer $26$ (indexed at $9$) of LLaMA3.1-8B on Hinting (IIT 4.0) (Sub-fig (d) in Appendix Fig. 12)
    \item Layer $29$ (indexed at $10$) of LLaMA3.1-8B on Strange Stories (2 scores) (IIT 4.0) (Sub-fig (d) in Appendix Fig. 12)
\end{enumerate}

There is no such case found for ToM Task with 3 scores.

\textbf{Criterion 2}: All four cases above exhibit significant effects (the definition of this significance can be found in Sec. \ref{Methods_IIT_for_LLM_Representations_Systematic_Spatio_temporal_Permutation}) as determined by Wilcoxon two-way tests for $\Phi$ values across different score categories, as shown in Appendix Fig. 6.

All potential layers suggesting "consciousness" phenomena originate from deeper layers. This is expected, as these layers are semantically rich, often encoding task-specific or context-dependent meanings. These layers play a crucial role in comprehending the overall meaning of sentences and in producing task-oriented outputs, such as predictions or classifications. High-level concepts, such as sentence-level sentiment, entailment, and other semantic phenomena, are strongly represented in these layers \citep{raganato2018analysis, jawahar-etal-2019-bert, sajjad2022analyzing}.

The case with Item $1$ above, which performs well under both IIT 3.0 and 4.0, corresponds approximately to the $\frac{2}{3}$ layer of LLaMA3.1-8B (the exact $\frac{2}{3}$ layer from our sampling procedure is Layer $21$ (indexed at $7$); see Sec. \ref{Methods_IIT_for_LLM_Representations_Representations_from_Multiple_Transformer_Layers} and Appendix A.1 for details). This finding is consistent with previous studies, which suggest that the $\frac{2}{3}$ layer is optimal for predicting human brain activity \citep{schrimpf2021neural, caucheteux2022brains, caucheteux2023evidence}.

\textbf{Criterion 3}: The comparisons of interpretation abilities across different metrics reveal that, in most cases, the Span Representation outperforms all estimates of consciousness from IIT. In only a few instances — namely, Layer 15 (indexed at 5) of Mistral-7B on False Belief and Layer 21 (the $\frac{2}{3}$ Layer) of LLaMA3.1-8B on False Belief — do all estimates for consciousness from IIT surpass the Span Representation with respect to the mean AUC. However, none of the four exceptional cases (Items $1$ to $3$, which satisfy both Criterion $1$ and Criterion $2$) exhibit compelling mean AUCs from any of the IIT estimates for consciousness.

In this regard, we are still cautious about claiming that any of the four cases described above are observed with "consciousness" phenomenon. This is because the different ToM Task test score categories are more likely to be attributed to the span-level information of the LLM representation sequence rather than to a "consciousness" phenomenon, as suggested by IIT estimates.

\subsection{Results Extended to Different Linguistic Spans}\label{Results_Extended_to_Different_Linguistic_Spans}

\footnote{All the figures in this section are provided in Appendix A.7.2.}Our study reveals quantitative distinctions in estimates of consciousness across different linguistic spans. Using the same settings as in Sec. \ref{Phi_Distribution_and_Comparison_across_ToM_Tasks_andLLMs}, with two additional filters — namely, the same Transformer Layer and the same Linguistic Span — Appendix Fig. 17 to Appendix Fig. 20 illustrate the distributions of $\Phi$ values across all linguistic spans (\textit{Entire}, \textit{Complement}, and \textit{MSV}). We limited the transformer layers to the last three layers and the $\frac{2}{3}$ layer for each LLM, as detailed in Sec. \ref{Methods_IIT_for_LLM_Representations_Representations_from_Multiple_Transformer_Layers}). Appendix Fig. 21 shows a similar analysis, except that the ToM Task (Strange Stories) uses 3 scores. The $\Phi$ value distribution, grouped according to $10$ randomization controls, is presented in Appendix Fig. 22, with Appendix Fig. 23 serving as the counterpart for the ToM Task (Strange Stories) with 3 scores. We further decompose the result shown in Appendix Fig. 22 into Appendix Fig. 24 and Appendix Fig. 25. The "good" and "bad" statistics are defined as in Sec. \ref{Phi_Distribution_and_Comparison_across_ToM_Tasks_andLLMs}. The analysis for the ToM Task with 3 scores is shown in Appendix Fig. 26. Finally, the interpretation abilities across different metrics, including the Span Representation, which is independent of estimating consciousness, are shown in Appendix Fig. 27 (and Appendix Fig. 28 for the ToM Task with 3 scores).

\textbf{Criterion $1$}: The following cases, although rare, show a significant predominance of "good" cases over "bad" ones. (The definition of significance is consistent with that described in Sec. \ref{Results_Extended_to_Different_Transformer_Layers}.)

\begin{enumerate}
    \item Layer $26$ (indexed at $9$) of LLaMA3.1-8B on Hinting (IIT 4.0) with Linguistic Span: \textit{Entire} (Sub-fig (d) in Appendix Fig. 24)
    \item Layer $29$ (indexed at $10$) of LLaMA3.1-8B on Strange Stories (IIT 4.0) with Linguistic Span: \textit{Entire} (Sub-fig (d) in Appendix Fig. 24)
    \item Layer $80$ (indexed at $11$) of LLaMA3.1-70B on Strange Stories (IIT 4.0) with Linguistic Span: \textit{Complement} (Sub-fig (d) in Appendix Fig. 25)
    \item Layer $53$ (indexed at $\frac{2}{3}$) of LLaMA3.1-70B on Hinting (IIT 4.0) with Linguistic Span: \textit{MSV} (Sub-fig (d) in Appendix Fig. 25)
\end{enumerate}

There is no such case found for ToM Task with 3 scores.

\textbf{Criterion 2}: All four of the cases described above exhibit significant effects, as determined by Wilcoxon two-way tests on $\Phi$ values between different score categories. These results are presented in Appendix Fig. 18 and Appendix Fig. 20.

Language comprehension, particularly the ability to identify and understand complement syntax and mental (state) verbs in sentences, is crucial for the development of ToM in human childhood \citep{astington1999longitudinal, de2002complements, de2007interface, milligan2007language, de2014linguistic}. However, our results reveal that this ability is not significantly represented in LLMs. This finding highlights potential discrepancies between natural and artificial intelligence with respect to mind development and language comprehension mechanisms, warranting further investigation.

\textbf{Criterion 3}: Comparisons of interpretation abilities across different metrics show that, in most cases, the Span Representation outperforms all estimates of consciousness from IIT. Only in rare instances — specifically, Layer 21 (the $\frac{2}{3}$ layer) of Mistral-7B on False Belief with the Linguistic Span: \textit{Complement}, and Layer 21 (the $\frac{2}{3}$ layer) of LLaMA3.1-8B with the Linguistic Span: \textit{Entire} on False Belief — do all estimates of consciousness from IIT surpass the Span Representation in terms of mean AUC. However, none of the four outstanding cases (Items $1$ to $4$, which satisfy both Criterion $1$ and Criterion $2$) exhibit a compelling mean AUC compared to any of the IIT estimates of consciousness.

In this context, we remain hesitant to claim that any of the four cases presented above are observed with "consciousness" phenomenon, as the variations in ToM task test score categories are more likely attributed to the span-level information of the LLM representation sequence, rather than to a "consciousness" phenomenon as suggested by IIT estimates.

\subsection{Additional Results Based on Systematic Permutations (Spatio)}\label{Additional_Results_Based_on_Systematic_Permutations}

\footnote{All the figures in this section are provided in Appendix A.7.3.}Additional results based on \textit{spatio} permutations are presented in Appendix Fig. 29 to Appendix Fig. 54. These correspond directly to the results in Fig. \ref{main_result_fig_1} and Fig. \ref{main_result_fig_2}, as well as those from Appendix Fig. 5 to Appendix Fig. 28, with the key difference being that the permutation controls are limited to \textit{spatio} rather than \textit{temporal}. For brevity, we omit detailed descriptions, settings, and filtering criteria here, as they can be found in Sec. \ref{Phi_Distribution_and_Comparison_across_ToM_Tasks_andLLMs} to Sec. \ref{Results_Extended_to_Different_Linguistic_Spans}. We focus only on the outstanding cases that are comparable to those discussed in the previous sections. These results provide contrasts and consolidations to the findings presented in Sec. \ref{Phi_Distribution_and_Comparison_across_ToM_Tasks_andLLMs}, Sec. \ref{Comparing_Metrics_Estimating_Consciousness_and_Other_Independent_Metric}, Sec. \ref{Results_Extended_to_Different_Transformer_Layers}, and Sec. \ref{Results_Extended_to_Different_Linguistic_Spans}.

In general, for the results: (\textbf{Criterion 1}) Only one outstanding case is identified, as shown in sub-fig (l) of Appendix Fig. 29, where the number of "good" cases significantly outweighs the "bad" ones. This case corresponds to Mixtral-8x7B on Strange Stories (2 scores) (IIT 4.0). No such case is observed for the ToM Task with 3 scores. (\textbf{Criterion 2}) The degree of effect, i.e., the significance ($p$ value) of the Wilcoxon two-way tests for $\Phi$ values between different score categories, aligns with expectations for this case, as shown in sub-fig (c) of Appendix Fig. 29.

\textbf{Criterion 3}: In this control group, the Span Representation does not consistently outperform the estimates for consciousness from IIT. There are several instances, such as LLaMA3.1-70B on Hinting and Irony, Mistral-7B on Irony, and LLaMA3.1-8B on Strange Stories (2 scores), where all the estimates for consciousness from IIT surpass the Span Representation in terms of mean AUC. Moreover, the Span Representation is not the top performer when compared to $\mu[\Phi]$ (IIT 4.0) in terms of mean AUC for the outstanding case (satisfying both Criterion 1 and 2), as shown in sub-fig (d) of Appendix Fig. 30. Additionally, there are other cases where the Span Representation loses its superiority.

For the comparison of different transformer layers, (\textbf{Criterion 1}) the following cases, though still rare, demonstrate a significant predominance of "good" cases over "bad" ones:

\begin{enumerate}
    \item Layer $15$ (indexed at $5$) of LLaMA3.1-8B on Strange Stories (IIT 3.0) (Sub-fig (c) in Appendix Fig. 38)
    \item Layer $24$ and $21$ (indexed at $8$ and $\frac{2}{3}$) of Mistral-7B on Irony (IIT 4.0) (Sub-fig (b) in Appendix Fig. 38)
    \item Layer $32$ (indexed at $11$) of Mixtral-8x7B on Strange Stories (2 scores) (IIT 4.0) (Sub-fig (b) in Appendix Fig. 39)
\end{enumerate}

There is no such case found for ToM Task with 3 scores.

\textbf{Criterion 2}: The degree of effect, represented by the significance ($p$ value) from Wilcoxon two-way tests for $\Phi$ values between different score categories, is not consistently satisfactory. The two cases corresponding to Item $1$ and Item $3$ are significant, as shown in Appendix Fig. 32 and Appendix Fig. 33, but the cases for Item $2$ do not reach significance, as shown in Appendix Fig. 31.

Once again, the majority of the potential layers that may indicate "consciousness" phenomena are derived from deeper layers and the $\frac{2}{3}$ layer of LLMs, as observed even with the \textit{spatio} permutation applied to the ARR.

\textbf{Criterion 3}: The Span Representation no longer consistently outperforms the estimates for consciousness from IIT. In fact, there are several instances where the mean AUC for the Span Representation ranks lowest, as well as other cases where it loses its superiority, as shown in Appendix Fig. 41 and Appendix Fig. 42. It is also noteworthy that in the identified outstanding cases, all estimates for consciousness from IIT surpass the Span Representation in terms of mean AUC (i.e., Item $2$), and the estimate for $\mu[\Phi]$ (IIT 4.0) surpasses the Span Representation for Item $3$. Notably, Item $3$ is the only case that satisfies all three criteria defined in Sec. \ref{Methods_Statistical_Analysis}.

For the results comparing different linguistic spans, (\textbf{Criterion $1$}) The following cases, although still rare, show a significant predominance of "good" cases over "bad" ones:

\begin{enumerate}
    \item Layer $32$ (indexed at $11$) of Mixtral-8x7B on Strange Stories (IIT 3.0) with Linguistic Span: \textit{Complement} (Sub-fig (a) in Appendix Fig. 51)
    \item Layer $21$ (indexed at $\frac{2}{3}$) of LLaMA3.1-8B on Hinting (IIT 3.0) with Linguistic Span: \textit{MSV} (Sub-fig (c) in Appendix Fig. 50)
    \item Layer $32$ (indexed at $11$) of Mixtral-8x7B on Strange Stories (IIT 4.0) with Linguistic Spans: \textit{Entire} and \textit{Complement} (Sub-fig (b) in Appendix Fig. 51)
    \item Layer $21$ (indexed at $\frac{2}{3}$) of Mistral-7B on Irony (IIT 4.0) with Linguistic Span: \textit{Entire} (Sub-fig (b) in Appendix Fig. 50)
\end{enumerate}

There is no such case found for ToM Task with 3 scores.

\textbf{Criterion 2}: The degree of effect, as indicated by the significance ($p$ value) of the Wilcoxon two-way tests for $\Phi$ values across different score categories, is not entirely satisfactory in this control group. The three cases for Items $2$ and $3$, as shown in Appendix Fig. 44 and Appendix Fig. 45, are satisfactory. However, the two cases for Items $1$ and $4$, as shown in Appendix Fig. 45 and Appendix Fig. 43, do not show statistical significance.

We observe no significant differences when examining various linguistic spans in comparison to the entire spectrum of each stimulus, even when analyzed using the \textit{spatio} permutation control on the CARR.

\textbf{Criterion 3}: The Span Representation no longer consistently outperforms the estimates for consciousness from IIT. In several cases, the mean AUC for the Span Representation ranks among the lowest, and in others, it loses its leading position, as shown in Appendix Fig. 41 and Appendix Fig. 42. Notably, for the identified outstanding cases, the estimates for $\mu[\Phi]$ (IIT 4.0) and $\mu[\Phi\text{-structures}]$ (IIT 4.0) outperform the Span Representation for Item $3$. In contrast, the mean AUCs for $\mu[\Phi^{\max}]$ (IIT 3.0) and $\mu[CI]$ (IIT 3.0) are comparable to those of the Span Representation for Item $1$. Additionally, the estimates for $\mu[\Phi]$ (IIT 4.0) and $\mu[\Phi\text{-structures}]$ (IIT 4.0) surpass the Span Representation for Item $4$. Notably, Item $3$ is associated with two additional examples that meet all criteria for identifying "consciousness" phenomena, as defined in Sec. \ref{Methods_Statistical_Analysis}.

\section{Discussion}\label{Discussion}

Although feed-forward systems, such as artificial neural networks, are generally considered incapable of consciousness according to IIT \citep{oizumi2014phenomenology, albantakis2023integrated}, their internal states — specifically, the sequences of representations generated by LLMs that encode their "knowledge," "understanding," "value," or "position" in the world — remain largely unexplored due to the black-box nature of deep learning. While LLM representations are being increasingly studied as analogs of neuroimaging data when comparing and aligning human brain activity with LLM "brain" activity in natural language processing \citep{schrimpf2021neural, caucheteux2022brains, caucheteux2023evidence, karamolegkou-etal-2023-mapping}, we pose a further inquiry: \textit{Is "experience" encoded in these sequences of representations beyond mere "knowledge," "understanding," "value," or "position"?} In this context, our primary focus has been directed toward understanding these representations with our primary research question: \textbf{Can "consciousness" be observed in the internal states of an LLM, specifically in its learned representations, particularly when analyzed as a sequence?} In essence, LLM-generated representations are multidimensional abstractions that encode lexical, syntactic, and semantic properties of text, enabling the model to reason, infer, and generate language with high proficiency. Variations in these representations across different LLMs reflect differences in how each model processes, understands, and encodes language. These discrepancies arise from the unique architectural designs, training data, and methodological choices inherent to each model.

We decomposed our primary research question and established three stringent criteria to identify cases with potential indications of consciousness underlying the (C)ARR, but after conducting large-scale experiments — restricted by \textit{temporal} permutation control — no cases meeting all three criteria were found among the valid samples. However, we identified several promising cases that satisfied two out of the three criteria, particularly in analyses across different transformer layers, as detailed in Sec. \ref{Results_Extended_to_Different_Transformer_Layers}. Notably, all these cases originated from deeper layers, aligning with existing knowledge that these layers primarily encode semantic information, often capturing task-specific or context-dependent meanings. Such layers are essential for comprehending the overall meaning of sentences and generating task-oriented outputs, including predictions and classifications. High-level concepts, such as sentence-level sentiment, entailment, and other semantic phenomena, are strongly encoded in these layers \citep{raganato2018analysis, jawahar-etal-2019-bert, sajjad2022analyzing}. Additionally, we observed that the case associated with approximately the $\frac{2}{3}$ layer belongs to this group (Sec. \ref{Results_Extended_to_Different_Transformer_Layers}), as this layer has been reported to exhibit strong alignment with human brain activity \citep{schrimpf2021neural, caucheteux2022brains, caucheteux2023evidence}.

Across various linguistic spans, we found no single case satisfying all three established criteria, though intriguing instances emerged that met two of the three criteria (detailed in Sec. \ref{Results_Extended_to_Different_Linguistic_Spans}). Critically, no substantive evidence suggested that directing response representations' attention to designated linguistic spans — specifically complement syntax and/or mental (state) verbs — would yield significant alterations. This finding is particularly noteworthy when contextualized against the established developmental psychology literature, which emphasizes language comprehension — particularly the identification and understanding of complement syntax and/or mental (state) verbs — as a fundamental mechanism in human childhood ToM development \citep{astington1999longitudinal, de2002complements, de2007interface, milligan2007language, de2014linguistic}. Our research reveals potential fundamental discrepancies between natural and artificial intelligence regarding mind development and language comprehension mechanisms, warranting further in-depth investigation.

Typically, arbitrarily permuting embeddings derived from LLMs is inadvisable, as these representations constitute critical learned features acquired during model training. The dimensional order within these vector representations is paramount, preserving semantic integrity and ensuring architectural compatibility with subsequent computational processing. Nevertheless, our study necessitates permutation along the embedding dimension for each (C)ARR, predicated on treating learned feature vectors as nodes within our proposed RN. We hypothesize potential latent interconnections (edges) between individual nodes or node clusters as inspired by \cite{nemirovsky2023implementation}. Consequently, \textit{spatio} permutations enable systematic exploration of these potential nodal relationships within the RN.

The findings from \textit{spatio} permutation controls complement those from \textit{temporal} ones, confirming that potential "consciousness" phenomena are associated with deeper layers, as well as the $\frac{2}{3}$ layer, in LLM representations. Furthermore, the CARR show insensitivity when directed to focus more on specific linguistic spans — abilities that are crucial for the development of ToM in human childhood. However, two cases were identified that met all three criteria, contrasting with the results from the \textit{temporal} permutation controls, which yielded no such cases. This suggests a potentially profound, yet tentative, indication of consciousness embedded within LLM representations, i.e.,

\begin{enumerate}
    \item Layer $32$ (indexed at $11$) of Mixtral-8x7B on Strange Stories (2 scores) (IIT 4.0) with Linguistic Spans: \textit{Entire} and \textit{Complement} (Sub-fig (b) in Appendix Fig. 51)
\end{enumerate}

Moreover, under \textit{spatio} permutation controls, numerous cases emerge wherein IIT consciousness estimates outperform Span Representation, demonstrating superior interpretative capabilities across diverse metrics relative to ToM test score categories. In several instances, Span Representation distinctly loses its previously dominant position (see Appendix Fig. 30, Appendix Fig. 41, Appendix Fig. 42, Appendix Fig. 53, and Appendix Fig. 54 for references). Conversely, \textit{temporal} permutation controls predominantly reveal Span Representation's superiority over consciousness metrics (see Fig. \ref{main_result_fig_2}, Appendix Fig. 15, Appendix Fig. 16, Appendix Fig. 27, and Appendix Fig. 28 for detailed comparisons).

The interpretation ability, quantified by mean AUC across diverse metrics relative to ToM test score categories, merits our critical attention. These metrics serve as independent indicators, deliberately decoupled from IIT assumptions — notably, the premise that a higher $\Phi$ value is presumed to exhibit a more conscious experience \citep{oizumi2014phenomenology, albantakis2023integrated}. Instances where IIT consciousness estimates surpass Span Representation potentially signify emergent possibilities of observed "consciousness" phenomenon emerged from LLM representations.

Recent studies \citep{brown2020language, wei2022emergent, bubeck2023sparks} have highlighted emergent abilities in LLMs, such as behaviors and problem-solving capabilities that are evident in larger models but absent or less pronounced in smaller ones. These abilities are particularly notable in domains such as in-context learning, complex reasoning, few-shot learning, and task generalization \citep{brown2020language, wei2022emergent, bubeck2023sparks}. Contrary to potential expectations, our investigation revealed no significant disparity in potential consciousness indicators between larger models (Mixtral-8x7B and LLaMA3.1-70B) and smaller counterparts (Mistral-7B and LLaMA3.1-8B). However, this finding should be interpreted with considerable caution due to three critical methodological limitations: 1) substantial sample loss during network initialization using PyPhi for IIT 4.0, particularly with the Mixtral-8x7B model; 2) reliance on quantization techniques for inference, necessitated by hardware constraints and detailed in Appendix A.1, which potentially compromises representation resolution and precision; and 3) dimensional reduction via PCA to manage computational complexity when calculating $\Phi$ values and associated IIT metrics.

While our study employed the IIT as the theory of consciousness, future research necessitates a comprehensive exploration of alternative theoretical frameworks for consciousness. IIT remains contentiously debated, with scholars critically challenging the validity of its axioms regarding consciousness phenomenology \citep{cerullo2015problem, doerig2019unfolding}. The theory's scientific standing is scrutinized by a collective statement from $124$ scholars arguing for its classification as pseudoscience \citep{fleming2023integrated}. Another widely discussed framework, Global Workspace Theory (GWT) emerges as a prominent one, conceptualizing conscious percepts as attentional spotlights mediated by executive control functions \citep{baars2017global, baars2021global}. Additionally, a diverse landscape of consciousness theories — including Recurrent Processing Theory, Computational Higher-order Theory, Attention Scheme Theory, Predictive Processing, and Agency and Embodiment — offer alternative investigative lenses for examining consciousness within AI systems, as comprehensively surveyed by \cite{butlin2023consciousness}.

Another limitation of our study is that it focuses solely on textual data, whereas contemporary AI systems are increasingly capable of processing multimodal data, including text, audio, and images etc. This is particularly evident in the field of agent-based multimodal intelligence, as discussed in \cite{durante2024agent}. We consider the exploration of multimodal as well as agentic AI systems producing/consuming the representations that build on top of LLMs as promising directions for future research.

\section{Conclusion}\label{Conclusion}

In summary, our study provides insights based on large-scale quantitative results into observing "consciousness" phenomenon from contemporary Transformer-based LLM representations, and confirms that we do not yet observe statistically significant signs of consciousness from the representations derived from these models according to the three stringent criteria. However, it is plausible that ongoing developments in LLMs may lead to models capable of generating representations being observed with "consciousness" phenomenon, as suggested by the findings from \textit{spatio} permutations, which warrant our further attention. Our carefully designed, model-agnostic approach marks the inauguration of such exploration, and offers a standardized framework for observing "consciousness" phenomenon in LLM representations.

\section{Data and Code Availability}\label{Data and Code Availability}

The original ToM test results were released along with \cite{strachan2024testing}, and the corresponding dataset is publicly accessible at \url{https://osf.io/dbn92}. The code and the processed intermediate data, i.e., the labeled linguistic spans along with their contexts for each response, and the augmented responses for each stimulus are available as the Supplementary Material at: \url{https://doi.org/10.1016/j.nlp.2025.100163}. Please refer to the "README.md" file in "Data and Code" for details.

\section{Declaration of Competing Interest}\label{Declaration of Competing Interest}

The authors declare that they have no known competing financial interests or personal relationships that could have appeared to influence the work reported in this paper.

\section{Declaration of Generative AI and AI-assisted Technologies in This Study}\label{Declaration of Generative AI and AI-assisted Technologies in This Study}

During the preparation of this work the author(s) used the web interfaces of:

\begin{itemize}
    \item GPT-4o, \url{https://chatgpt.com/}, and the specific version we used is gpt-4o-2024-08-06.
    \item Claude 3.5 Sonnet, \url{https://claude.ai/}, and the specific version we used is claude-3-5-sonnet@20240620. And,
    \item Gemini, \url{https://gemini.google.com/}, and the specific version we used is google/gemini-1.5-flash-002.
\end{itemize}

in order to 

\begin{itemize}
    \item facilitate our data processing w.r.t labeling different linguistic features in stimuli (detailed in Sec. \ref{Methods_IIT_for_LLM_Representations_Response_Representations_Attended_to_Stimulus_Linguistic_Spans_in_Context}), and
    \item text augmentation for responses (detailed in Sec. \ref{Methods_IIT_for_LLM_Representations_Text_Augmentation_for_Responses}), and
    \item to improve the readability and language of this work
\end{itemize}

After using this tool/service, the author(s) reviewed and edited the content as needed and take(s) full responsibility for the content of the publication.

\section{Acknowledgments}\label{Acknowledgments}

We appreciate Mr. Aryaman Reddi, currently a PhD candidate at TU Darmstadt, Germany, since we had some inspiring discussions at the early stage of the project.

\bibliographystyle{cas-model2-names}

\bibliography{cas-refs}

\end{document}